\documentclass[letterpaper]{article} 
\usepackage{aaai23}  
\usepackage{times}  
\usepackage{helvet}  
\usepackage{courier}  
\usepackage[hyphens]{url}  
\usepackage{graphicx} 
\usepackage{natbib}  
\usepackage{caption} 
\usepackage{algorithm}
\usepackage{algorithmic}

\usepackage{amsmath}
\graphicspath{ {./fig/} }
\usepackage{subfigure}
\usepackage{amsmath} 
\usepackage{bm}      

\usepackage{newfloat}
\usepackage{listings}
\DeclareCaptionStyle{ruled}{labelfont=normalfont,labelsep=colon,strut=off} 
\floatstyle{ruled}
\newfloat{listing}{tb}{lst}{}
\floatname{listing}{Listing}
\title{Aesthetically Relevant Image Captioning}

\author{
    Zhipeng Zhong\textsuperscript{\rm{1, 3, 4, 5}},  
    Fei Zhou\textsuperscript{\rm{1, 2, 3, 4, 5}} and  
    Guoping Qiu\textsuperscript{\rm{1, 2, 3, 4, 5, 6}}
}

\affiliations{

    \textsuperscript{\rm 1}College of Electronics and Information Engineering, Shenzhen University, China, \textsuperscript{\rm 2}Peng Cheng National Laboratory, Shenzhen, China,  \textsuperscript{\rm 3}Guangdong Key Laboratory of Intelligent Information Processing, Shenzhen, China,  \textsuperscript{\rm 4}Shenzhen Institute for Artificial Intelligence and Robotics for Society, China, \textsuperscript{\rm 5}Guangdong-Hong Kong Joint Laboratory for Big Data Imaging and Communication, Shenzhen, China and \textsuperscript{\rm 6}School of Computer Science, The University of Nottingham, UK

}

\usepackage{bibentry}

\begin{document}

\maketitle

\begin{abstract}
Image aesthetic quality assessment (AQA) aims to assign numerical aesthetic ratings to images whilst image aesthetic captioning (IAC) aims to generate textual descriptions of the aesthetic aspects of images. 
In this paper, we study image AQA and IAC together and present a new IAC method termed Aesthetically Relevant Image Captioning (ARIC). Based on the observation that most textual comments of an image are about objects and their interactions rather than aspects of aesthetics, we first introduce the concept of Aesthetic Relevance Score (ARS) of a sentence and have developed a model to automatically label a sentence with its ARS. We then use the ARS to design the ARIC model which includes an ARS weighted IAC loss function and an ARS based diverse aesthetic caption selector (DACS). We present extensive experimental results to show the soundness of the ARS concept and the effectiveness of the ARIC model by demonstrating that texts with higher ARS's can predict the aesthetic ratings more accurately and that the new ARIC model can generate more accurate, aesthetically more relevant and more diverse image captions. Furthermore, a large new research database containing $510K$ images with over $5$ million comments and $350K$ aesthetic scores, and code for implementing ARIC are available at \url{https://github.com/PengZai/ARIC}.  
\end{abstract}

\section{Introduction}

Image aesthetic quality assessment (AQA) aims to automatically score the aesthetic values of images. This is very challenging because aesthetics is a highly subjective concept. 
AQA models are either regressors or classifiers that extract image features and output the aesthetic scores or classes \cite{AIA2021Survey}. Many images, especially those on photography competition websites contain both aesthetic scores and textual comments. It has been shown that including both the visual and textual information can improve AQA performances \cite{multimodalAQA-2016, Zhang2021MSCANMS}. 

Unlike visual contents which are rather abstract, texts are much easier for human to comprehend. Image captioning, which aims to automatically generate textual descriptions of images has been extensively researched, and much progress has been achieved in recent years with the help of deep learning technology \cite{DeepLearningImageCaptioning2019}. Whilst the vast majority of researchers have focused on image captions about objects and their relations and interactions, the more abstract and arguably more challenging problem of image aesthetic captioning (IAC) \cite{chang2017aesthetic}, which aims to generate comments about the aesthetic aspects of images, has received much less attention.

In the existing literature, image AQA and IAC are studied independent from each other. However, these are closely related areas of aesthetic visual computing, we therefore believe that jointly study them is beneficial. For example, currently IAC performances are evaluated either subjectively or based on metrics such as SPICE \cite{anderson2016spice} and BLEU \cite{papineni2002bleu} which evaluate the similarity between the generated and reference (ground truth) sentences rather than the aesthetic relevance of the texts. It will be very useful if we can directly measure the aesthetic relevance of the IAC results, for example, how accurate the generated caption can predict the images aesthetic scores. 


In this paper, we first contribute a large research database called DPC2022 which contains 510K images, over 5 million comments and 350K aesthetic ratings. We then present a new IAC method called Aesthetically Relevant Image Captioning (ARIC). Based on the observation that most image comments are general descriptions of image contents and not about their aesthetics, we first introduce the concept of Aesthetic Relevance Score (ARS) of a sentence. ARS consists of 5 components including scores related to aesthetic words, the length of the sentence, words describing objects, the sentiments of the sentence, and term frequency-inverse document frequency (tf-idf). A list of aesthetic and object words has been manually constructed from DPC2022. After (automatically) labelling the comments in DPC2022 with the ARS scores, we then construct an ARS predictor based on the Bidirectional Encoder Representations (BERT) language representation model \cite{devlin2018bert}. 

The introduction of ARS and its predictor have enabled the design of the ARIC model which includes an ARS weighted IAC loss function and an ARS based diverse aesthetic caption selector (DACS). Unlike methods in the literature that simply learned a direct mapping between the images and their comments in the database, regardless of the aesthetic relevance of the comments (in fact many of the texts have nothing to do with aesthetics),  the ARIC model is constructed based on ARS weighted loss function which ensures that it learns aesthetically relevant information. Furthermore, unlike traditional methods that pick the output sentences based on the generator's confidence which is not directly based on aesthetic relevance, the introduction of the ARS has enabled the design of DACS which can output a diverse set of aesthetically highly relevant sentences. In addition, we have fine tuned the powerful pre-trained image and text matching model CLIP \cite{radford2021learning} using DPC2022 as an alternative to ARS for selecting aesthetically relevant captions. 

We have performed extensive experiments. As DPC2022 is by far one of the largest AQA databases, we first provide baseline AQA results. We then present ARIC's image aesthetic captioning performances to demonstrate its effectiveness.  In summary, the contributions of this paper are
\begin{enumerate}
	\item A large image database, DPC2022, is constructed for researching image aesthetic captioning and image aesthetic quality assessment. DPC2022 is the largest dataset containing both aesthetic comments and scores. 
	\item A new concept, aesthetic relevance score (ARS), is introduced to measure the aesthetic relevance of sentences. Lists of key words and other statistical information for constructing the ARS model are made available. 
	\item Based on ARS, we have developed the new aesthetically relevant image captioning (ARIC) system capable of producing not only aesthetically relevant but also diverse image captions. 
\end{enumerate}

\section{Related Work }
\textbf{Image aesthetic quality assessment (IAQA)}. One of the main challenges in AQA is the lack of large scale high quality annotated datasets. The aesthetic visual analysis (AVA) dataset~\cite{murray2012ava} contains $250K$ images, each has an aesthetic score, and other labels. 
This is still  one of the most widely used databases in aesthetic quality assessment. The aesthetic ratings from online data (AROD) dataset~\cite{schwarz2018will} contains $380K$ images from a photo sharing website. The score of an image is based on the number of viewers who liked it, which can be an unreliable indicator. 
Recent AQA systems are mostly based on deep learning neural networks and supervised learning that take the images or their textual descriptions or both as input to predict the aesthetic scores \cite{valenzise2022advances, zhang2020beyond}. 

\textbf{Image aesthetic captioning (IAC)}. Aesthetic image captioning was first proposed in \cite{chang2017aesthetic} where the authors also presented the photo critique captioning dataset (PCCD) which contains pair-wise image comment data from professional photographers. It contains 4235 images and more than sixty thousands captions. The AVA-Captions dataset \cite{ghosal2019aesthetic} was obtained by using a probabilistic caption filtering method to clean the noise of the original AVA captions. It has about $230K$ images with roughly 5 captions per image. The DPC-Captions dataset \cite{jin2019aesthetic} contains over $150K$ images and nearly $2.5$ million comments where each comment was automatically annotated with one of the 5 aesthetic attributes of the PCCD through knowledge transfer. Very recently, the Reddit Photo Critique Dataset (RPCD) was published by \cite{RPCD}. This dataset  contains tuples of image and photo critiques which has $74K$ images and $220K$ comments. Image aesthetic captioning is an under explored area and existing works mostly used LSTM model to generate aesthetic captions. 


\textbf{Image captioning}. Image captioning aims to generate syntactically and semantically correct sentences to describe images. This is a complex and challenging task in which many deep learning-based techniques have been developed in recent years \cite{DeepLearningImageCaptioning2019}.  Training deep models requires large amount of annotated data which are very difficult to obtain. VisualGPT \cite{Chen_2022_CVPR} leverages the linguistic knowledge from a large pre-trained language model GPT-2 \cite{radford2019language} and quickly adapt it to the new domain of image captioning. It has been shown that this Transformer \cite{Transformer} based technique has superior performances to LSTM based methods used in previous IAC works \cite{chang2017aesthetic}\cite{ghosal2019aesthetic}\cite{jin2019aesthetic}.  We adopt this method for IAC to provide benchmark performances for the newly established DPC2022 dataset.      

\section{The DPC2022 Dataset }
\begin{figure}[htbp]
    \centering
    \subfigure[]{
    \includegraphics[width=3.8cm]{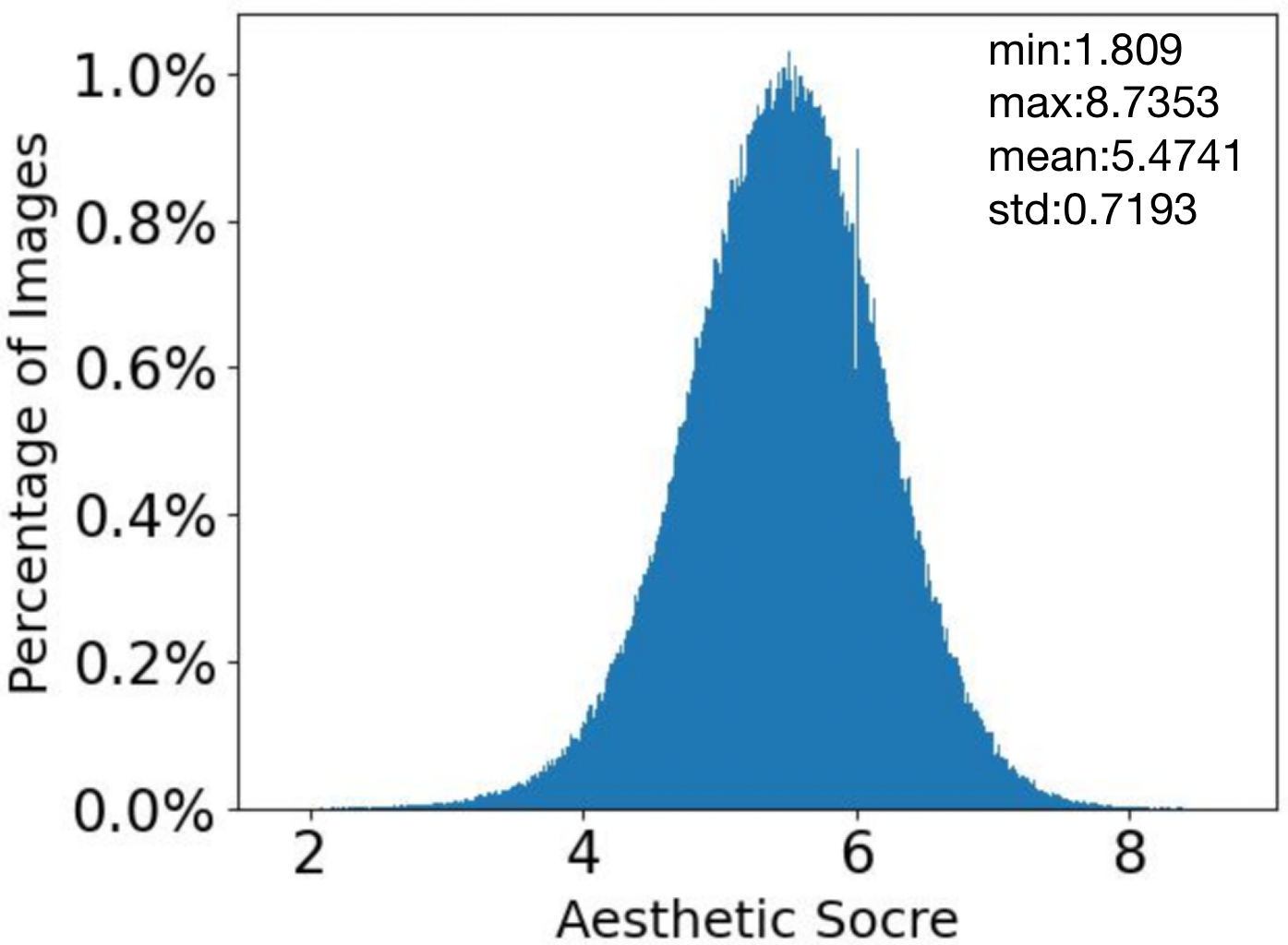}
    }
    \quad
    \subfigure[]{
	\includegraphics[width=3.8cm]{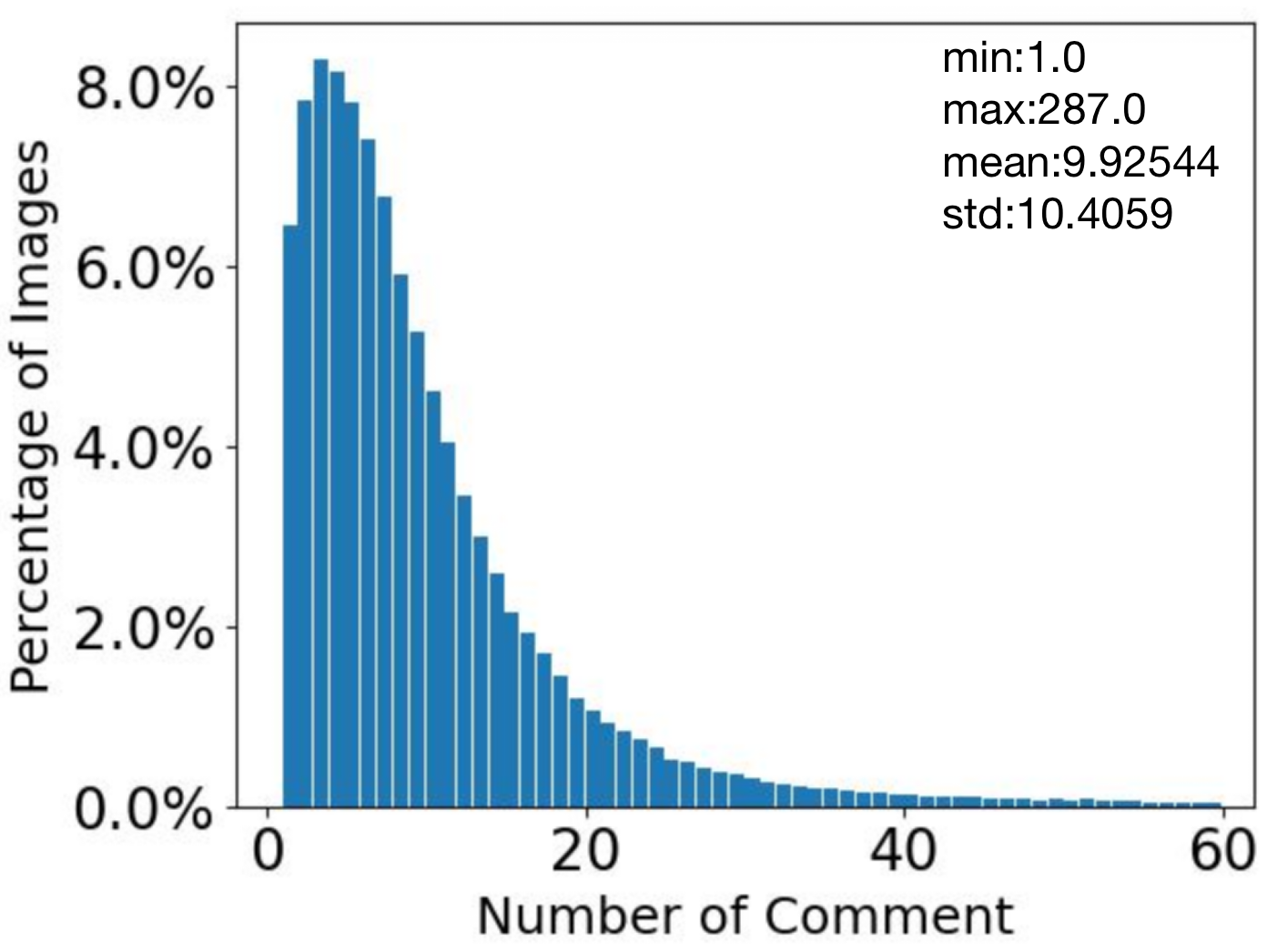}
    }
    \quad    
    \subfigure[]{
    \includegraphics[width=3.8cm]{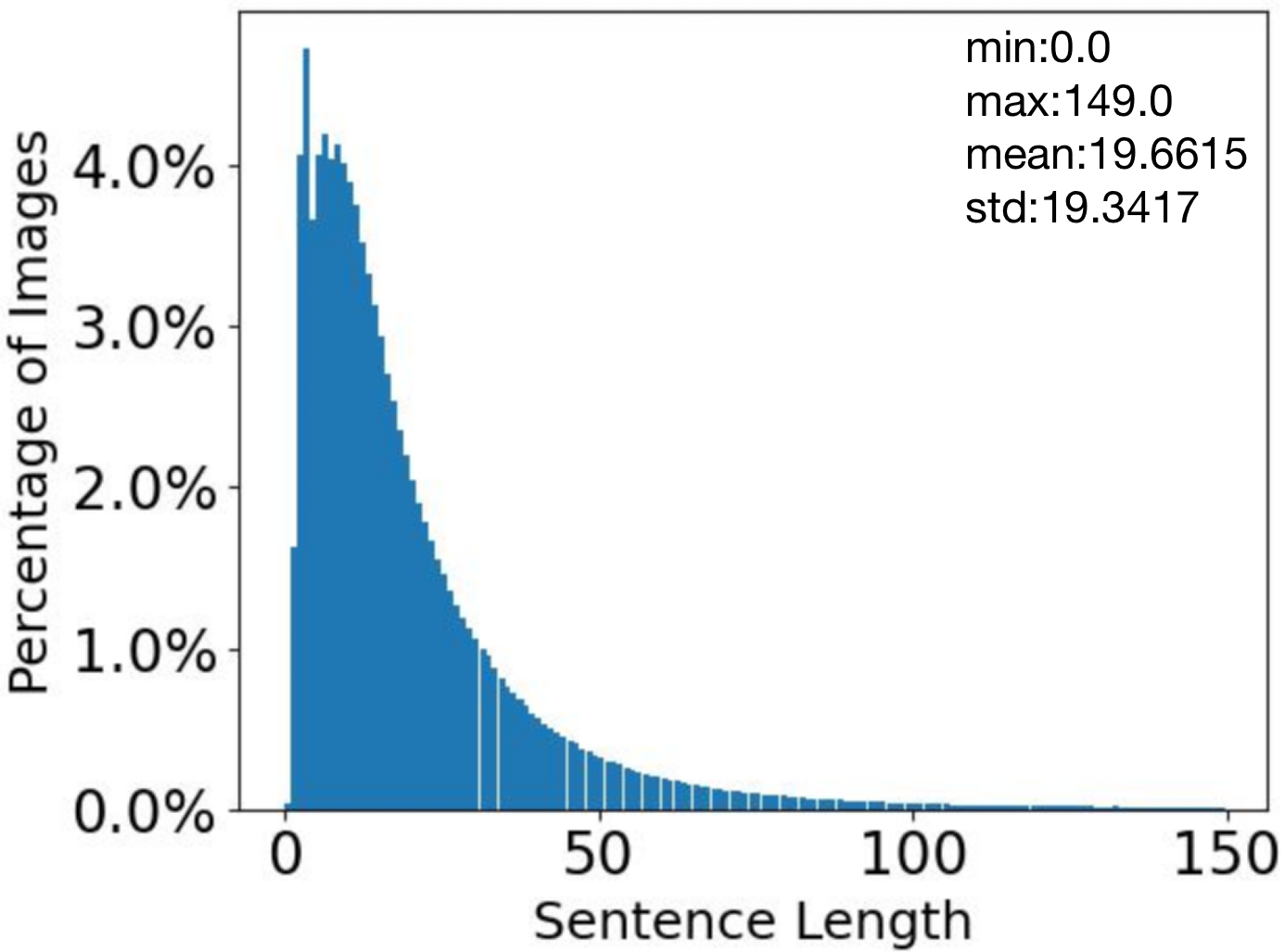}
    }
    \quad
    \subfigure[]{
	\includegraphics[width=3.8cm]{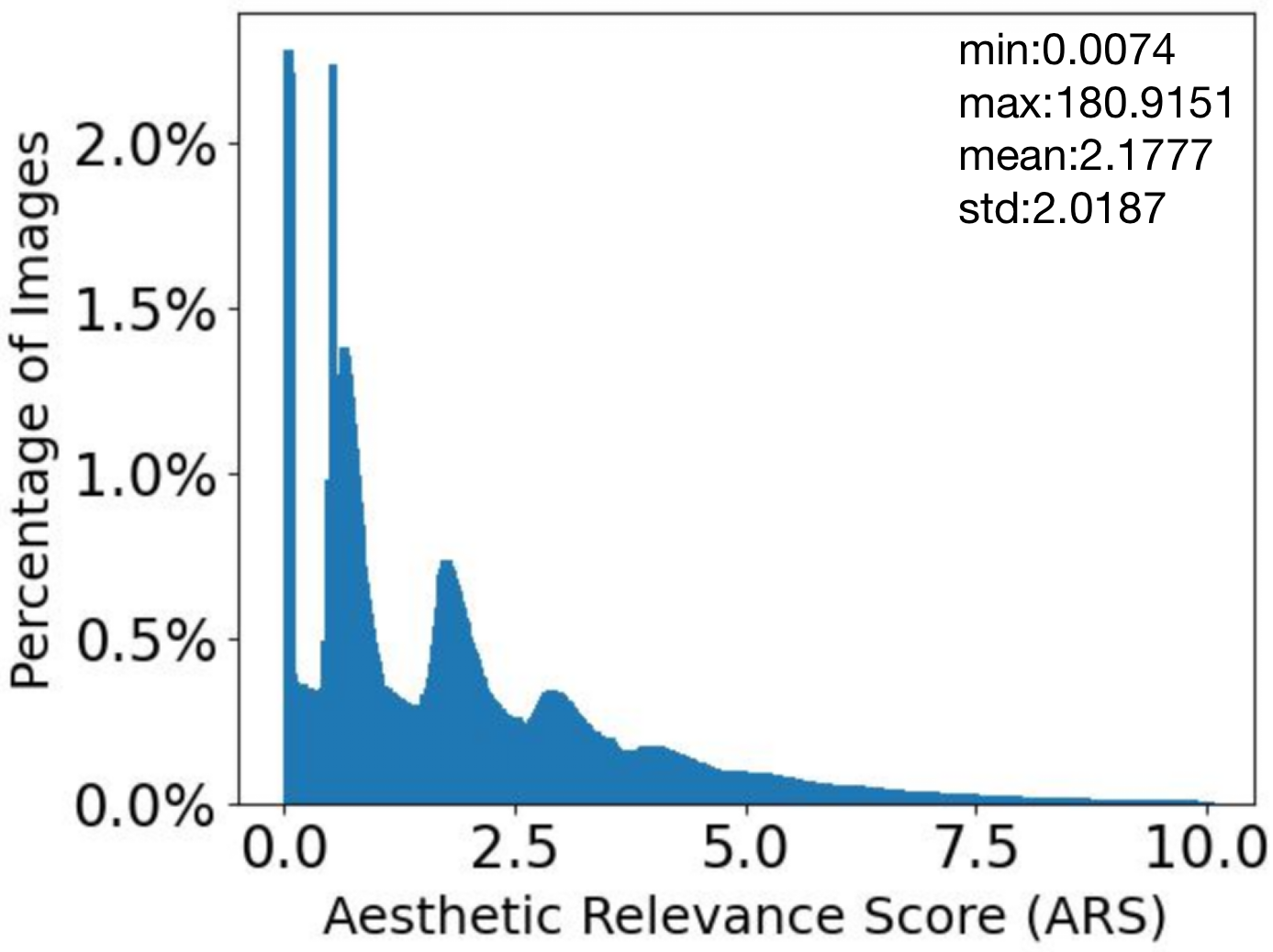}
    }
    \caption{Basic statistics of the DPC2022 dataset.}
    \label{fig.1}
\end{figure}

The AVA dataset~\cite{murray2012ava} which was constructed a decade ago remains to be one of the largest and most widely used in image AQA. As discussed above, datasets available for IAC are either small or constructed from the original AVA dataset. In the past 10 years, the source website of the AVA dataset (www.dpchallenge.com) has been continuously organising more photography competitions and has accumulated a lot more photos and comments. We therefore believe it is a good time to make full use of the data currently available from the website to construct a new dataset to advance research in image aesthetic computing. 
We first crawled the website and grabbed all currently available images and their comments. Initially we obtain a total of $780K$ images and their comments. We then used an industrial strength natural language processing tool \textit{spaCy} (https://spacy.io/) to clean the data by removing items such as emoji and other strange spellings, symbols and punctuation marks. At the end, we have kept $510K$ images with good quality and clean comments. Within these $510K$ images, there are $350K$ with both comments and aesthetic scores ranging between 1 and 10.  Figure \ref{fig.1} shows the statistics of the DPC2022 dataset. On average, each image contains roughly 10 comments, each comments on average contains 21 sentences, and the average sentence length is 19 words.

\section{Image Aesthetic Relevance of Text} 

The dictionary definition of aesthetic is ``concerned with beauty or the appreciation of beauty", this is a vague and subjective concept. Photographers often use visual characteristics such as lighting, composition, subject matter and colour schemes to create aesthetic photos. Comments associated with images, especially those on the internet websites, can refer to a variety of topics, not all words are relevant to the aesthetics of images, and some words may be more relevant than others. To the best of the authors' knowledge, there exist no definitions of image aesthetic relevant words and phrases in the computational aesthetic literature. And yet when inspecting the comments from the DPC2022 dataset, it is clear that many have nothing to do with the images' aesthetic qualities. It is therefore appropriate to distinguish words that refer to the aesthetic quality and those that are not relevant. We have developed the Aesthetic Relevant Score (ARS)  to quantify a comment's aesthetic relevance. It is based on a mixture of subjective judgement and statistics from the dataset. Before describing the ARS in detail, it is appropriate to note that this is by no means the only way to define such a quantity, 
however, we will demonstrate its usefulness through application in image aesthetic captioning and multi-modal image aesthetic quality assessment. 

\subsection{Labelling a Sentence with its ARS}
\label{section:datalabel}
The \textit{ARS} of a sentence $t$ is defined as: 
 \begin{equation}
 \label{ARD_definition}
 ARS (t) = A(t)+L(t)+ O(t) + S(t)  + T_{fidf}(t)
 \end{equation}
 
where $A(t)$ is related to aesthetic words, $L(t)$ is related to the length of $t$, $O(t)$ is related to object words, $S(t)$ is the sentiment score of $t$, and $T_{fidf}(t)$ is related to term frequency–inverse document frequency. 

The components in (\ref{ARD_definition}) are computed based on statistics of the DPC2022 database and details of the computational procedures are in \textbf{Appendix I} in the supplementary materials. 
For computing $A(t)$, we manually selected 1022 most frequently appeared image aesthetics related words such as \textit{shot, color, composition, light, focus, background, subject, detail, contrast, etc.}, the full list of these words, $\{AW_{list}\}$, can be found in \textbf{Appendix II}. For computing $O(t)$, we manually selected 2146 words related to objects such as \textit{eye, sky, face, ribbon, water, tree, flower, expression, hand, bird, glass, dog, hair, cat, smile, sun, window, car, etc}, the full list of these words, $\{OW_{list}\}$, can be found in \textbf{Appendix III}. The sentiment score $S(t)$ is calculated based on the $BerTweet$ model \cite{perez2021pysentimiento}. Figure \ref{fig.1}(d) shows the distribution of \textit{ARS} of the DPC2022 dataset. It is seen that many sentences contain no aesthetic relevant information, and the majority of the comments contain very low aesthetic relevant information. Informally inspecting the data shows that this is reasonable and expected.

\begin{figure}[th]
	\centering
	\includegraphics[width=0.7\columnwidth]{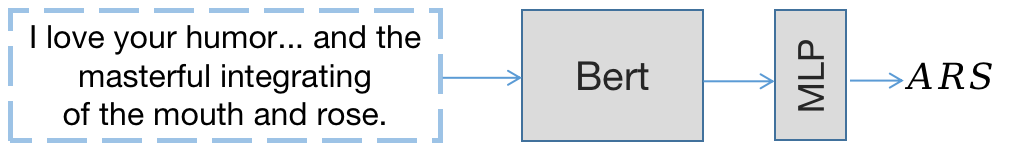} 
	\caption{ARS Predictor}.
	\label{fig-ARDPredictor}
\end{figure}

\subsection{Automatically Predicting the ARS}

For the \textit{ARS} to be useful, we need to be able to predict any given text's aesthetic relevant score. With the labelled data described above, we adopt the  pre-trained Bidirectional Encoder Representations (BERT) language representation model \cite{devlin2018bert} for this purpose. A $768\times1$ fully connected layer is cascaded to the output of BERT 
as shown in Figure \ref{fig-ARDPredictor}. We train the model with the mean squared error (MSE) loss function. Table \ref{tab:ARSresult} shows the \textit{ARS} prediction performance. It is seen that the Spearman's rank-order correlation (SRCC) and the Pearson linear correlation coefficient (PLCC) are both above 0.95, indicating excellent prediction accuracy. It is therefore possible to predict a sentence's ARS with an architecture shown in Figure \ref{fig-ARDPredictor}.

\begin{table}[th]
	\begin{center}
		\resizebox{\linewidth}{!}{
             \begin{tabular}{|c|c|c|c|c|}
				\hline
				Metrics    & SRCC$\uparrow$   & PLCC$\uparrow$   & RMSE$\downarrow$    & MAE$\downarrow$  \\ \hline
				Performance  & 0.9553 & 0.9599 & 0.5617  & 0.3395 \\ \hline

		\end{tabular}}
		\caption{ARS prediction accuracy.} 
		\label{tab:ARSresult}	
	\end{center}
\end{table}

\section{Aesthetically Relevant Image Captioning }


\subsection{Image Captioning Model}

\begin{figure}[th]
	\centering
	\includegraphics[width=1\columnwidth]{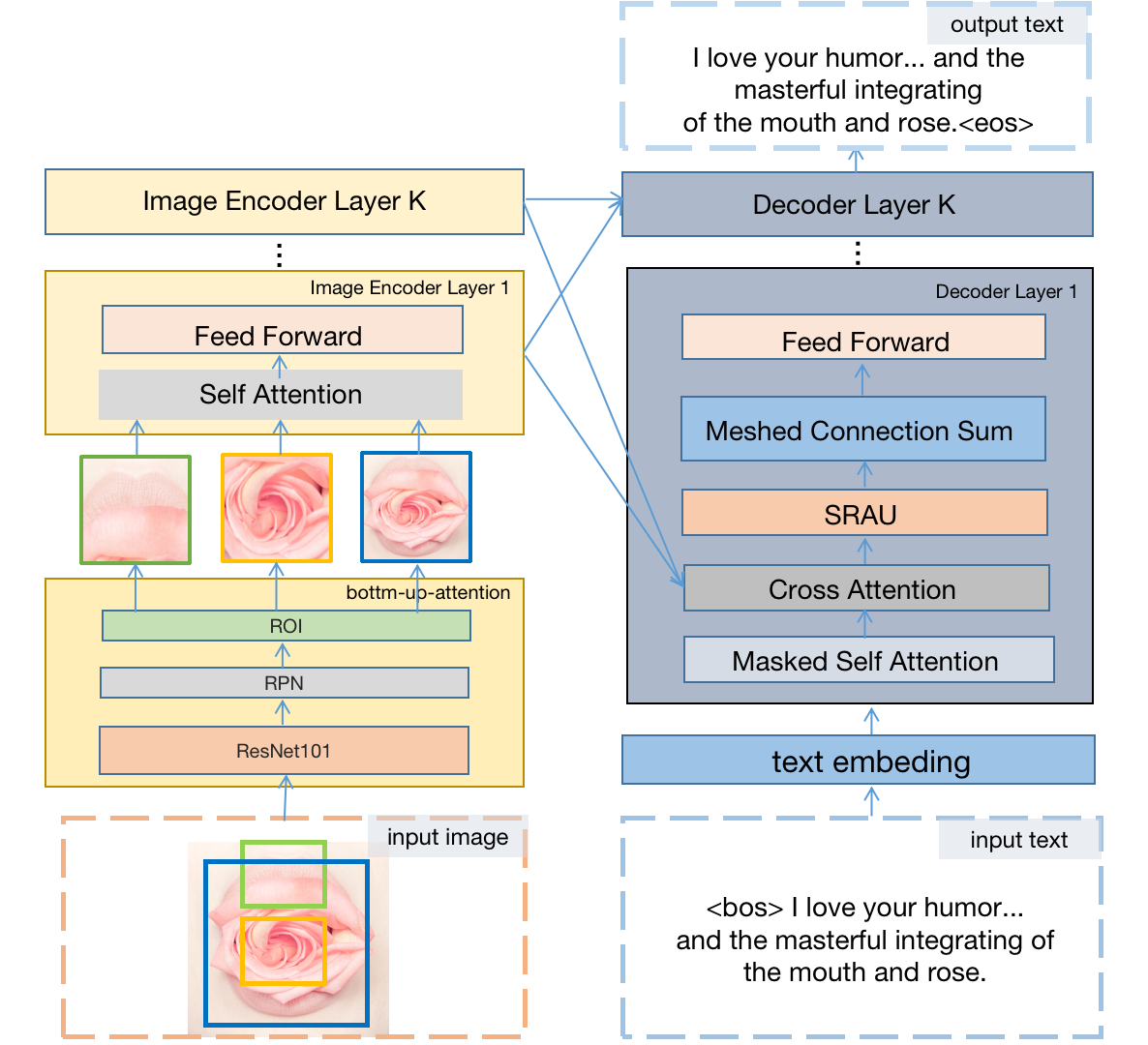} 
	\caption{Image captioning model. Left: object detection is performed on the input image, visual features and positions of the regions of interest are encoded. Right: image caption is generated based on the encoded image features. Note the input text is paired with the input image for training purpose.}
	\label{fig-captionmodel}
\end{figure}

For image captioning, many models based on LSTM \cite{LSTM-Captioning} and Transformer \cite{Transformer} have been developed. Given an image, we first extract visual features using an encoder structure, 
then use a decoder to generate image captions as shown in Figure \ref{fig-captionmodel}. To obtain image embedding, we follow the bottom-up-attention model \cite{anderson2018bottom} and use ResNet-101 \cite{he2016deep} as the backbone network of the Faster R-CNN \cite{ren2015faster} to extract image features. The bottom-up attention model uses a region proposal network (RPN) and a region-of-interest (ROI) pooling strategy for object detection. In this paper, we retain 50 most interesting regions and pass them to the encoder for image caption generation. Each region is represented by a 2048-dimensional feature vector.  

For the decoder, we use VisualGPT \cite{Chen_2022_CVPR} to generate image aesthetic captions. VisualGPT is an encoder-decoder transformer structure based on GPT-2 \cite{radford2019language} which is a powerful language model but does not have the ability to understand images. VisualGPT uses a self-resurrecting activation unit to encode visual information into the language model, and balances the visual encoder module and the language deocder module. 

\subsection{Aesthetically Relevant Loss Function}
 Unlike general image captioning which in most cases is about revealing the content and semantics of the image, e.g., objects in the image and their relations, image aesthetic captioning (IAC) should focus on learning aesthetically relevant aspects of the image. Arguably, IAC is much more challenging than generic image captioning. The very few existing IAC related literature, e.g., \cite{chang2017aesthetic}, \cite{jin2019aesthetic} and \cite{ghosal2019aesthetic}, simply treat the raw image comments from the training data as the IAC ground truth. However, as discussed previously, many of the texts contain no or very low aesthetic relevant information. The \textit{ARS}($t$) quantitatively measures the aesthetic relevance of a piece of text $t$, a higher \textit{ARS}($t$) indicates that $t$ contains high aesthetic relevant information and a low \textit{ARS}($t$) indicates otherwise. We therefore use \textit{ARS}($t$) to define the aesthetically relevant loss function to construct the IAC model. 
 
 Given training set $\{T(k)\}=\{I, y^{*} (t_k)\}$ which contains pairs of input image $I$ and one of its corresponding ground-truth caption sentences $y^{*}(t_k)= (y_1^{*}(t_k), y_2^{*}(t_k), ..., y_{N_k}^{*}(t_k) )$ consisting of words $y_i^{*}(t_k)$,  $i=1, 2, ..., N_k$, we generate a caption sentence $y(t_k)= (y_1(t_k), y_2(t_k), ..., y_{N_k}(t_k) )$ that maximises the following \textit{ARS}($t$) weighted cross-entropy loss
 
 \begin{equation}
 	\label{WeightedLossFunction}
	L_{AR}(\theta)=-\sum_{k=1}^{k=|T|}ARS(t_k)\sum_{i=1}^{N_k}\mathrm{log} ~p\left(y_i(t_k)=y_i^{*} (t_k)| \bm{\theta}\right)
 \end{equation}
 
where $t_k$ is the $k^{th} $ training sentence (note one image $I$ can have multiple sentences), $|T|$ is the training set size (in terms of sentences), $N_k$ is the number of words in the $k^{th} $ training sentence, and $\bm{\theta}$ is the model parameters.

\subsection{Diverse Aesthetic Caption Selector (DACS)}
Traditional IAC models such as the one described in Figure \ref{fig-captionmodel} output sentences based on the generator's confidence. All existing methods in the literature  \cite{chang2017aesthetic}, \cite{jin2019aesthetic} and \cite{ghosal2019aesthetic} adopt this approach. However, the generator's confidence is not directly based on aesthetic relevance. The new ARS concept introduced in this paper has provided a tool to measure the aesthetic value of a sentence, which in turn can help selecting aesthetically more relevant and more diverse captions from the generator.

Based on the ARS score defined in (\ref{ARD_definition}) for a piece of text $t$, we design a sentence selector method that enables the IAC model to generate diverse and aesthetically relevant sentences. Given a picture, we use the IAC model and beam search \cite{NPLHandbook} to generate $N$ most confident sentences. Then we use a sentence transformer\footnote{https://huggingface.co/models, \textbf{all-miniLM-L6-v2}} to extract the features of the sentences, and then calculate the cosine similarity among the $N$ sentences. We group sentences whose cosine similarity is higher than 0.7 into the same group such that each group contains many similar sentences. Then, we use the \textit{ARS} predictor in Figure \ref{fig-ARDPredictor} to estimate the \textit{ARS} of the sentences in each group. If the average \textit{ARS} of a group's sentences is below the mean \textit{ARS} of the training data ($2.1787$), then the group is discarded because their aesthetic values are low. From the remaining groups, we then pick the sentence with the highest \textit{ARS} in each group as the output. With this method, which we call the Diverse Aesthetic Caption Selector (DACS), we generate aesthetically highly relevant and diverse image captions. In the experiment section, we will show that OpenAI's CLIP \cite{radford2021learning}, a powerful image and text embedding model that can be used to find the text snippet best represents a given image, can be fine tuned to play the role of ARS for picking the best sentence in a group as output. 


\section{Multi-modal Aesthetic Quality Assessment}

The purposes of performing multi-modal AQA are two folds. Firstly, the newly established DPC2022 is one of the largest publicly available datasets and quite unique in the sense that it contains both comments and aesthetic scores. We want to provide some AQA performance baselines. Secondly, we make the reasonable assumption that the more accurate a piece of text can predict an image's aesthetic rating, the more aesthetically relevant is the text to the image. 

\begin{figure}[th]
\centering
\includegraphics[width=0.8\columnwidth]{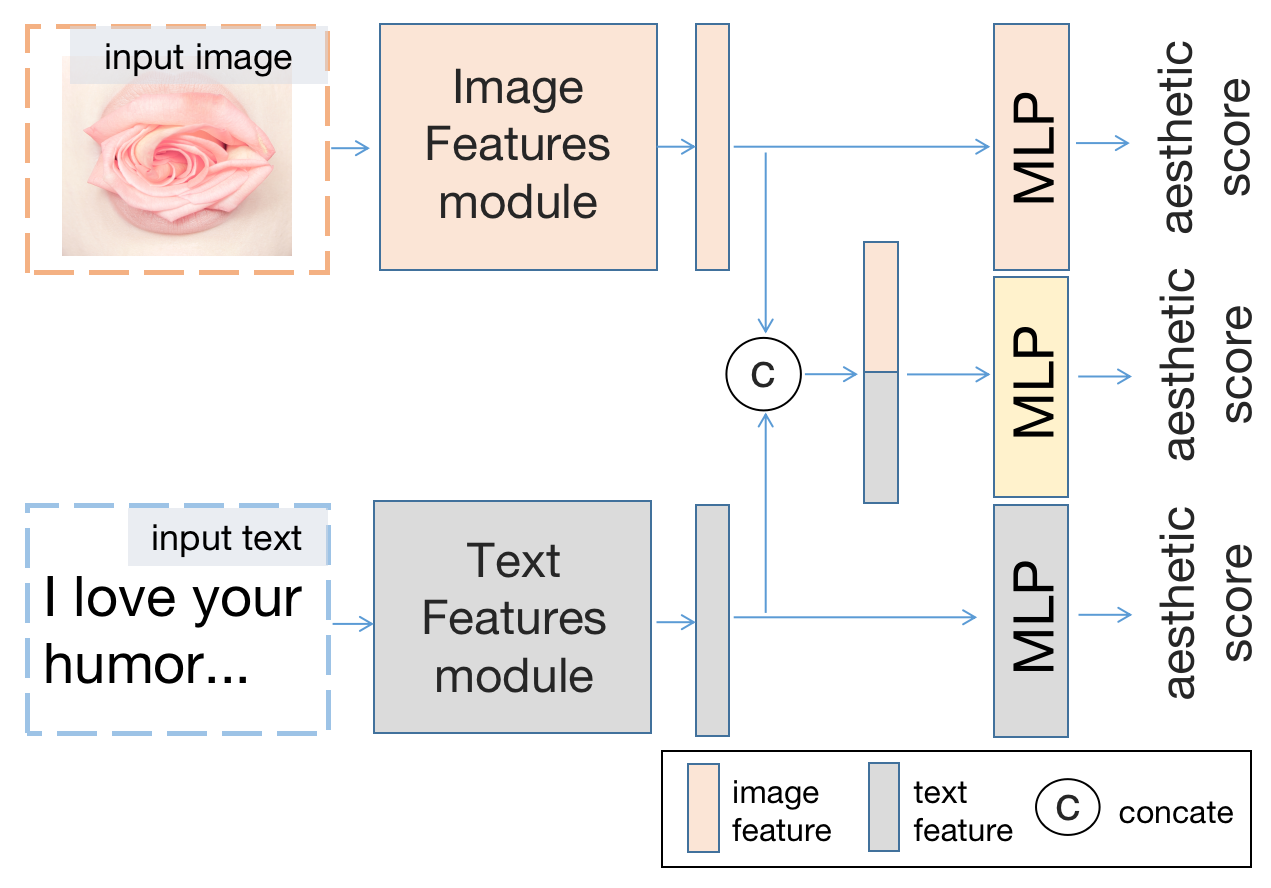} 
\caption{Multi-modal Image Aesthetic Quality Assessment.}
\label{fig-MMIAQA}
\end{figure}

The image AQA model is shown in Figure \ref{fig-MMIAQA}. There are 3 separate paths. The first takes the image as input and output the image's aesthetic rating. A backbone neural network is used for feature extraction. The features are then fed to an MLP to regress the aesthete rating. The second path takes the textual description of the image as input and output the image's aesthetic score. Again, a neural network backbone is used for textual feature extraction. The features are fed to an MLP network to output the aesthetic score. The third path is used to implement multi-modal AQA where visual and textual features are concatenated and fed to a MLP to output the aesthetic score. 

\section{Experimental Results}
We perform experiments based on the newly constructed DPC2022 dataset. It has a total of 510,000 photos where each photo also contains a review text. 350,000 out of the 510,000 photos also have aesthetic ratings. We call the dataset containing all images setA, and the subset containing images with both reviews and aesthetic ratings SetB. We use SetA for photo aesthetic captioning experiment and SetB for multi-modal aesthetic quality assessment. We divide SetB into a test set containing 106,971 images and a validation set containing 10,698 images, and the remaining 232,331 images are used as the training set. In photo aesthetic captioning, we use the same test set and validation set as those used in multi-modal aesthetic quality assessment, and the remaining SetA data (392,331 images) is used for training. 

\subsection{Evaluation Criteria}
Similar to those in the literature, we use 5 metrics to measure the performance of image AQA including Binary Accuracy (ACC), Spearman rank order correlation coefficient (SRCC), Pearson linear correlation coefficient (PLCC), root mean square error (RMSE), and mean absolute error(MAE). For image captioning, we use 4 metrics widely used in related literature including 
CIDER \cite{vedantam2015cider}, METETOR \cite{banerjee2005meteor}, ROUGE \cite{lin2004rouge}, and  SPICE \cite{anderson2016spice}. SPICE is a word-based semantic similarity measure for scene graphs, and the others are all based on n-gram.

\subsection{Implementation Details}
For image based AQA, we have used VGG-16, RESNET18, DESNET121, RESNEXT50, and ViT \cite{dosovitskiy2020image} as the backbone network and used their pre-trained models rather than training from scratch. The images are resized to 224x224, randomly flipped horizontally, and normalised to have a mean of 0.5 and a variance 0.5. For text based AQA, we have used the TEXTCNN \cite{kim-2014-convolutional}, TEXTRNN \cite{textrcnn}, BERT \cite{devlin2018bert} and ROBERTA \cite{liu2019roberta}) as the backbone network. For BERT and ROBERTA, we use their pre-trained models. Finally, we limit the number of tokens for each image's comment to 512. For IAC, we used the pre-trained GPT2 model and set token size to 64. All experiments were performed on a machine with 4 NVIDIA A100 GPUs. Adam optimizer with a learning rate of $2e^{-5}$ without weight decay was used. The AQA models were trained for 16 epochs and the IAC models were trained for 32 epochs. Codes will be made publicly available. 


\subsection{Image AQA Baseline Results}
In the first set of experiments, we experimented various network architectures in order to obtain some baseline image AQA results. Table \ref{tab:iaa experiment result} lists the baseline results of image based, text based, and mutlti-modal AQA results when the backbone used different network architectures. These results show that the textual reviews of the images are aesthetically highly relevant. In two out of 5 metrics, using the review text only gives the best performances. It is also seen that in 3 other metrics, multi-modal AQA has the best performances.     

\begin{table}[th]
	\begin{center}
		\resizebox{\linewidth}{!}{
			\begin{tabular}{|c|c|c|c|c|c|}
				\hline
				Backbone    & ACC$\uparrow$    & SRCC$\uparrow$   & PLCC$\uparrow$   & RMSE$\downarrow$    & MAE$\downarrow$  \\ \hline
				\multicolumn{6}{|c|}{Image based AQA} \\ 
				\hline
				VGG16       & 0.7848  & 0.5914 & 0.6078 & 0.5977  & 0.4699 \\ \hline
				RESNET18    & 0.8007 & 0.5946 & 0.6108 & 0.5850 & 0.4588  \\ \hline
				DESNET121 & 0.8096 & 0.6331 & 0.6471 & 0.5501 & 0.4287 \\ \hline
				RESNEXT50 & 0.8067 & 0.6386 & 0.6530 & 0.5486 & 0.4282 \\ \hline
				ViT & 0.8194 & 0.6755 & 0.6868 & 0.5356 & 0.4193 \\ \hline
				\multicolumn{6}{|c|}{Text based AQA} \\ 
				\hline
				TEXTCNN       & 0.812  & 0.6069 & 0.6251 & 0.461  & 0.5891 \\ \hline
				TEXTRCNN    & 0.8665 & 0.7516 & 0.7730 & 0.3611 & 0.4649 \\ \hline
				BERT & 0.8810 & 0.8024 & 0.8219 & 0.3292 & 0.4235 \\ \hline
				ROBERTA & \textbf{0.8939} & 0.8334 & 0.8551 & \textbf{0.2988} & 0.3826 \\ \hline
				\multicolumn{6}{|c|}{Multi-modal (image plus text) AQA} \\ 
				\hline
				RESNEXT50+RNN & 0.8591  & 0.8289 & 0.8456 & 0.3864  & 0.2998 \\ \hline
				RESNEXT50+BERT  & 0.8666 & 0.8493 & 0.8695 & 0.3780 & 0.2941 \\ \hline
				ViT+RNN & 0.8622 & 0.8386 & 0.8537 & 0.3747 & 0.2901 \\ \hline
				ViT+ROBERTA & 0.8693 & \textbf{0.8629} & \textbf{0.8803} & 0.3545 & \textbf{0.2756} \\ \hline
				ViT+BERT & 0.8697 & 0.8594 & 0.8766 & 0.3604 & 0.2814 \\ \hline
				
		\end{tabular}}
		\caption{AQA baseline results of DPC2022}
		\label{tab:iaa experiment result}	
	\end{center}
\end{table}

\begin{table}[h]
	\begin{center}
		\resizebox{\linewidth}{!}{
			\begin{tabular}{|c|c|c|c|c|c|}
				\hline
				Data Group    & ACC$\uparrow$    & SRCC$\uparrow$   & PLCC$\uparrow$   & RMSE$\downarrow$    & MAE $\downarrow$  \\ \hline
				All & 0.8666  & 0.8657 & 0.8814 & 0.3688 & 0.2883 \\ \hline
				Low $A$  & 0.8497 & 0.8247 & 0.8403 & 0.3725 & 0.2907 \\ \hline
				High $A$ & 0.8966 & 0.9150 & 0.9193 & 0.3621 & 0.2841 \\ \hline
				Low $O$  & 0.8516 & 0.8335 & 0.8498 & 0.3714 & 0.2903 \\ \hline		
				High $O$ & 0.8954 & 0.908 & 0.9127 & 0.3639 & 0.2845 \\ \hline		
				Low $L$  & 0.8521 & 0.8238 & 0.837 & 0.3704 & 0.2895 \\ \hline
				High $L$ & 0.8944 & \textbf{0.9184} & \textbf{0.9213} & 0.3658 & 0.2860 \\ \hline
				Low $S$  & 0.8358 & 0.814 & 0.8274 & 0.375 & 0.2937 \\ \hline
				High $S$ & \textbf{0.9300} & 0.9015 & 0.9144 & \textbf{0.3557} & \textbf{0.2773} \\ \hline
				Low $T_{fidf}$ & 0.8502 & 0.8182 & 0.8332 & 0.3714 & 0.2908 \\ \hline
				High $T_{fidf}$ & 0.8898 & 0.9107 & 0.9146 & 0.3651 & 0.2849\\ \hline
				Low $ARS$  & 0.8486 & 0.8216 & 0.8352 & 0.3726 & 0.2910 \\ \hline
				High $ARS$ & \underline{0.8991} & \underline{0.9157} & \underline{0.9202} & \underline{0.3620} & \underline{0.2835} \\ \hline
		\end{tabular}}
		\caption{Multi-modal AQA for images with high and low aesthetic relevant comments. These results are all based on VIT+BERT model.}
		\label{tab:comment information maa relationship}	
	\end{center}
\end{table}

\begin{table}[h]
	\begin{center}
		\resizebox{\linewidth}{!}{
			\begin{tabular}{|c|c|c|c|c|c|}
				\hline
				ARS Threshold    & ACC$\uparrow$    & SRCC$\uparrow$   & PLCC$\uparrow$   & RMSE$\downarrow$    & MAE $\downarrow$  \\ \hline
				$ARS \leq m_{ARS} -1.0\sigma_{ARS}$ & 0.8070 & 0.5992 & 0.6443 & 0.4252 & 0.3298 \\ \hline
				$ARS \leq m_{ARS} -0.8\sigma_{ARS}$  & 0.8213 & 0.6969 & 0.7172 & 0.3985 & 0.3079 \\ \hline
                $ARS \leq m_{ARS} -0.6\sigma_{ARS}$  & 0.8350 & 0.7602 & 0.7755 & 0.3848 & 0.3003 \\ \hline
                $ARS \leq m_{ARS} -0.4\sigma_{ARS}$  & 0.8415 & 0.7948 & 0.8091 & 0.3761 & 0.2940 \\ \hline
                $ARS \leq m_{ARS} -0.2\sigma_{ARS}$ & 0.8446 & 0.8105 & 0.8231 & 0.3750 & 0.2931 \\ \hline
                $ARS \leq m_{ARS} -0.0\sigma_{ARS}$ & 0.8486  & 0.8216 & 0.8352 & 0.3726 & 0.2910 \\ \hline
				$ARS \geq m_{ARS} +0.0\sigma_{ARS}$ & 0.8991 & 0.9157 & 0.9202 & 0.3620 &0.2835 \\ \hline
				$ARS \geq m_{ARS} +0.2\sigma_{ARS}$  & 0.9098 & 0.9206 & 0.9248 & 0.3626 & 0.2836 \\ \hline
				$ARS \geq m_{ARS} +0.4\sigma_{ARS}$  & 0.9129 & 0.9244 & 0.9261 & 0.3662 & 0.2863 \\ \hline
				$ARS \geq m_{ARS} +0.6\sigma_{ARS}$  & 0.9183 & 0.9274 & 0.9280 & 0.3652 & 0.2835\\ \hline
				$ARS \geq m_{ARS} +0.8\sigma_{ARS}$  & 0.9237 & 0.9280 & 0.9291 & 0.3670 & 0.2852 \\ \hline
				$ARS \geq m_{ARS} +1.0\sigma_{ARS}$  & 0.9232 & 0.9294 & 0.9291 & 0.3667 & 0.2854 \\ \hline
				ungroup(all)  & 0.8666 & 0.8657 &0.8814 & 0.3688 & 0.2883 \\ \hline
		\end{tabular}}
		\caption{Multi-modal image AQA. As the ARS of the comment texts increases, the aesthetic rating prediction performances increases. $m_{ARS}$ is the mean of \textit{ARS} and $\sigma_{ARS}$ is the variance of the \textit{ARS} of the DPC2022 dataset. $ARS \leq m_{ARS} - \alpha\sigma_{ARS}$ indicates the group of images each with a comment having an \textit{ARS} value less than or equal to $m_{ARS} - \alpha\sigma_{ARS}$.}
		\label{tab:comment information maa progressive relationship}	
	\end{center}
\end{table}

\subsection{Verification of ARS}

To verify the soundness of the newly introduced \textit{ARS}, we divide the DPC2022 validation set into two groups according to the ARS. 
If the \textit{ARS} model is sound, then we would expect a review text with a higher \textit{ARS} should be able to predict the image's aesthetic rating more accurately. Conversely, a review text that has a lower \textit{ARS} would produce a less accurate prediction of the image's aesthetic rating. 
Table \ref{tab:comment information maa relationship} shows the multi-modal AQA performances for two groups of testing samples. Low $ARS$ ($A, O, L, S, T_{fidf}$) represent the group where test samples have a $ARS$ ($A, O, L, S, T_{fidf}$) lower than their respective average values. High $ARS$ ($A, O, L, S, T_{fidf}$) represent the group where test samples have a $ARS$ ($A, O, L, S, T_{fidf}$) higher than their respective average values. Table 
\ref{tab:comment information maa progressive relationship} shows that as the \textit{ARS} increases, so does the AQA performances. These results clearly show that comments with high \textit{ARS} can consistently predict the images aesthetic rating more accurately than those with low \textit{ARS}. This means that \textit{ARS} and its components can indeed measure the aesthetic relevance of textual comments.

\begin{figure}[htbp]
    \centering
    \subfigure[]{
    \includegraphics[width=3.8cm]{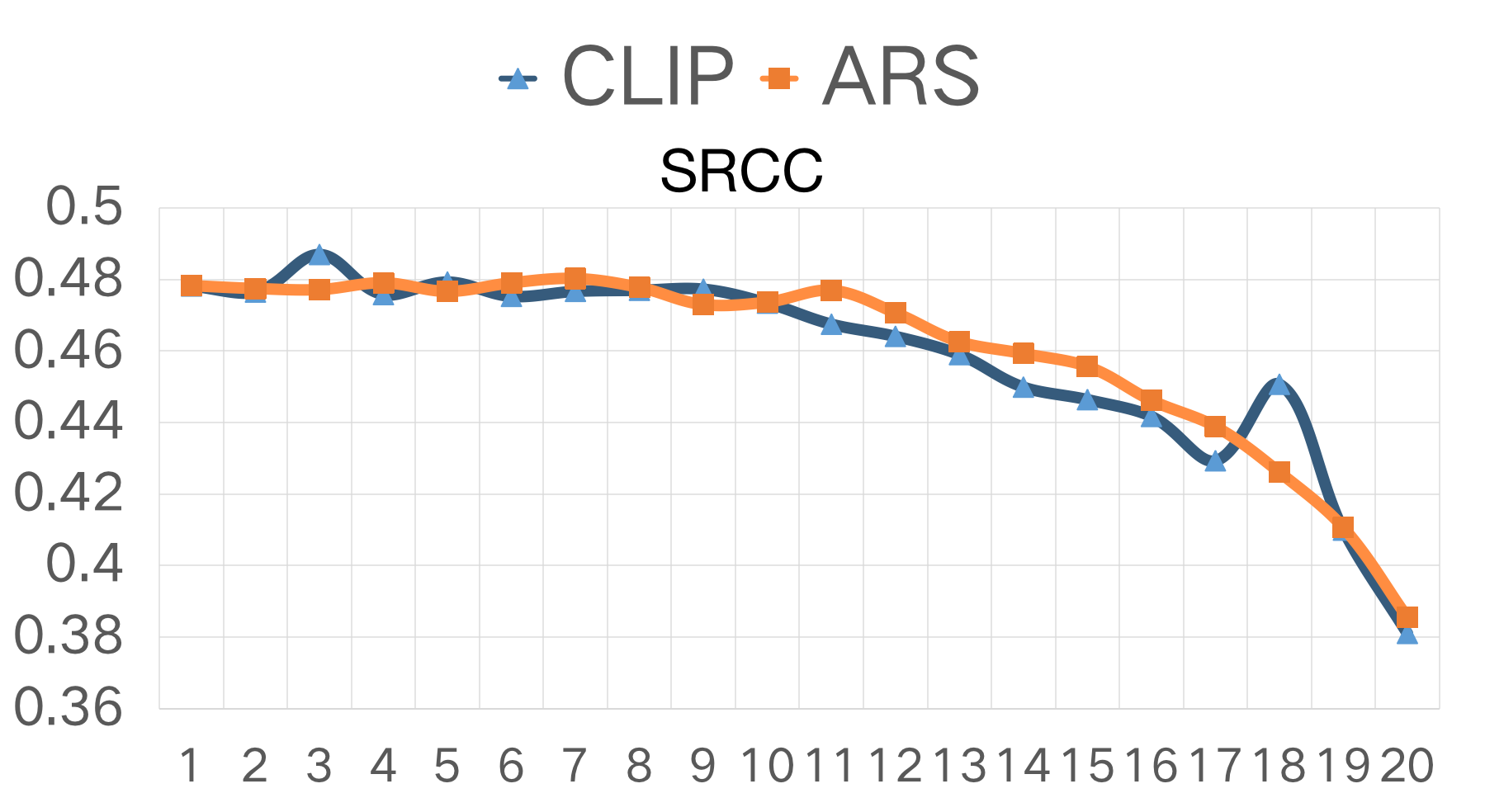}
    }
    \subfigure[]{
	\includegraphics[width=3.8cm]{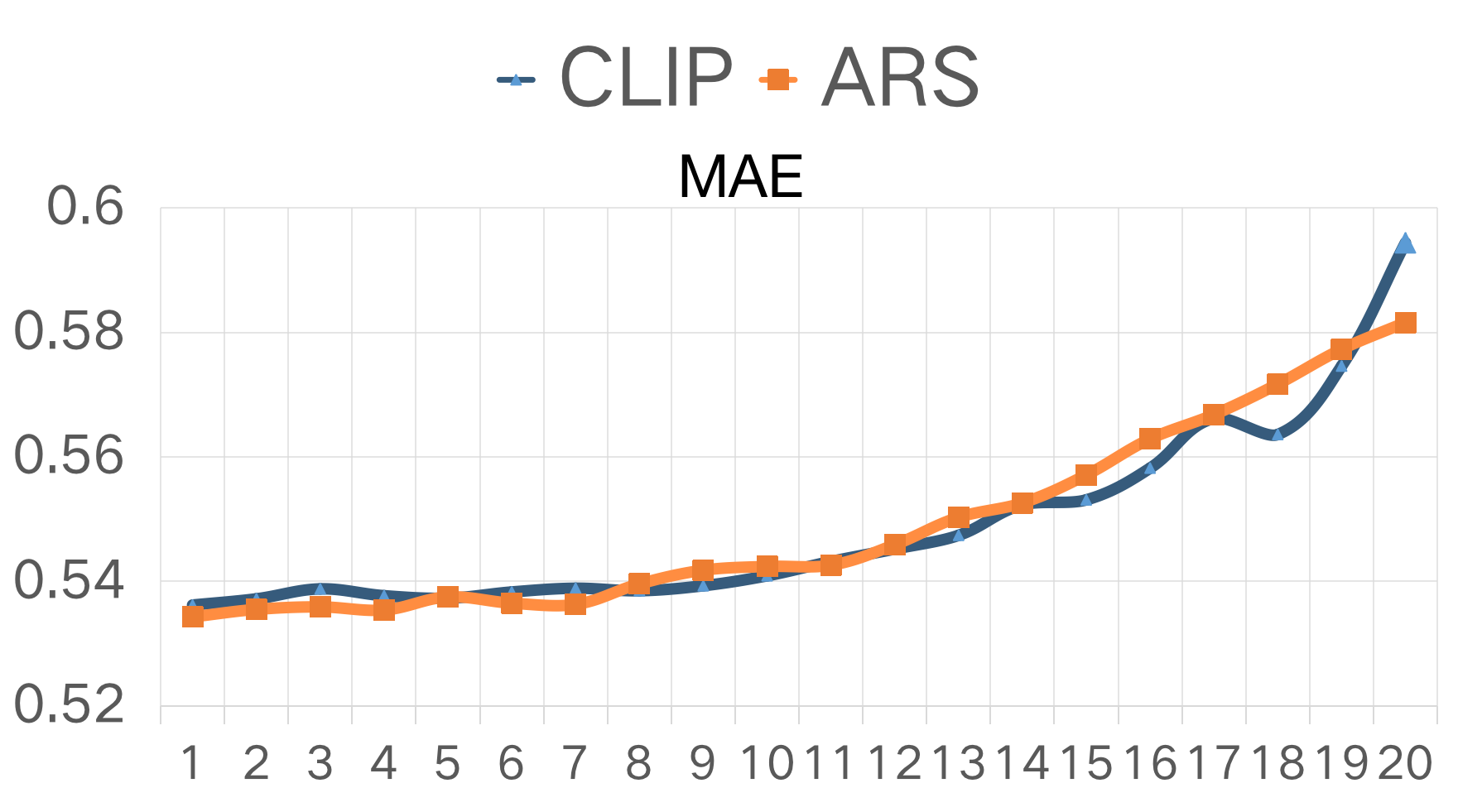}
    }
    \caption{Comparison between CLIP and ARS for ranking aesthetically relevant sentences for image AQA. Horizontal-axis is the rank position.}
    \label{CLIP-ARS}
\end{figure}

\subsection{ARS versus CLIP for Ranking Text Relevance}
CLIP (Contrastive Language-Image Pre-Training) is a neural network trained on a variety of (image, text) pairs \cite{radford2021learning}. It is a powerful model that can match natural language descriptions with visual contents. In this experiment, we first fine tune the pre-trained CLIP model with the training set of the DPC2022 data such that the images and their corresponding comments are matched. With such a fine tuned model, we can rank the sentences in the comments of the images. Let CLIP$(I, t)$ be the matching score between an image $I$ and a sentence $t$. Suppose we have two sentences $t_i$ and $t_j$, if CLIP$(I, t_i) >$ CLIP$(I, t_j)$ then $t_i$ and $I$ is a better match. Because CLIP has been fine tuned on DPC2022, it is reasonable to assume that if a text and an image has a better CLIP matching score, then the text can describe the image more accurately. Similar to \textit{ARS}, we can use CLIP to rank the sentences according to their CLIP matching scores. To compare the CLIP and \textit{ARS} in the selection of texts for image AQA, we rank the sentences of the images in the test set and perform text based 
AQA for the sentences in different ranks. The SRCC and MAE performances of those ranked by \textit{ARS} and CLIP are shown in Figure \ref{CLIP-ARS}. It is seen that the higher ranking sentences by both methods can predict the aesthetic rating more accurately, and that both seem to perform very similarly. These results show that the properly fine tuned CLIP model can also be used to select aesthetically relevant text for image AQA. It is worth noting that \textit{ARS} is a very simple scheme while CLIP is a much more complex but powerful model. Figures \ref{fig-diversityresult15} shows visual examples of how the sentences are ranked by \textit{ARS} and CLIP.  

\subsection{Aesthetic Captioning Performances}

Table \ref{tab:infromation comment score improve caption improvement.} shows the aesthetic image captioning performances of our new aesthetically relevant model (ARIC) based on the loss function $L_{AR}(\theta)$ as defined in (\ref{WeightedLossFunction}). For comparison, we have implemented a baseline model in which standard cross entropy loss function is used, i.e., setting $ARS(t_k)=1$ in (\ref{WeightedLossFunction}). 
It is seen that our new model consistently outperforms the baseline model. These results demonstrate the usefulness and effectiveness of introducing the aesthetically relevant score (\textit{ARS}) for aesthetics image captioning. Even though these metrics do not directly measure aesthetic relevance, the better performances of the new method nevertheless demonstrate the soundness of the new algorithm design. 

\begin{figure*}[h]
	\centering
	\includegraphics[width=0.9\textwidth]{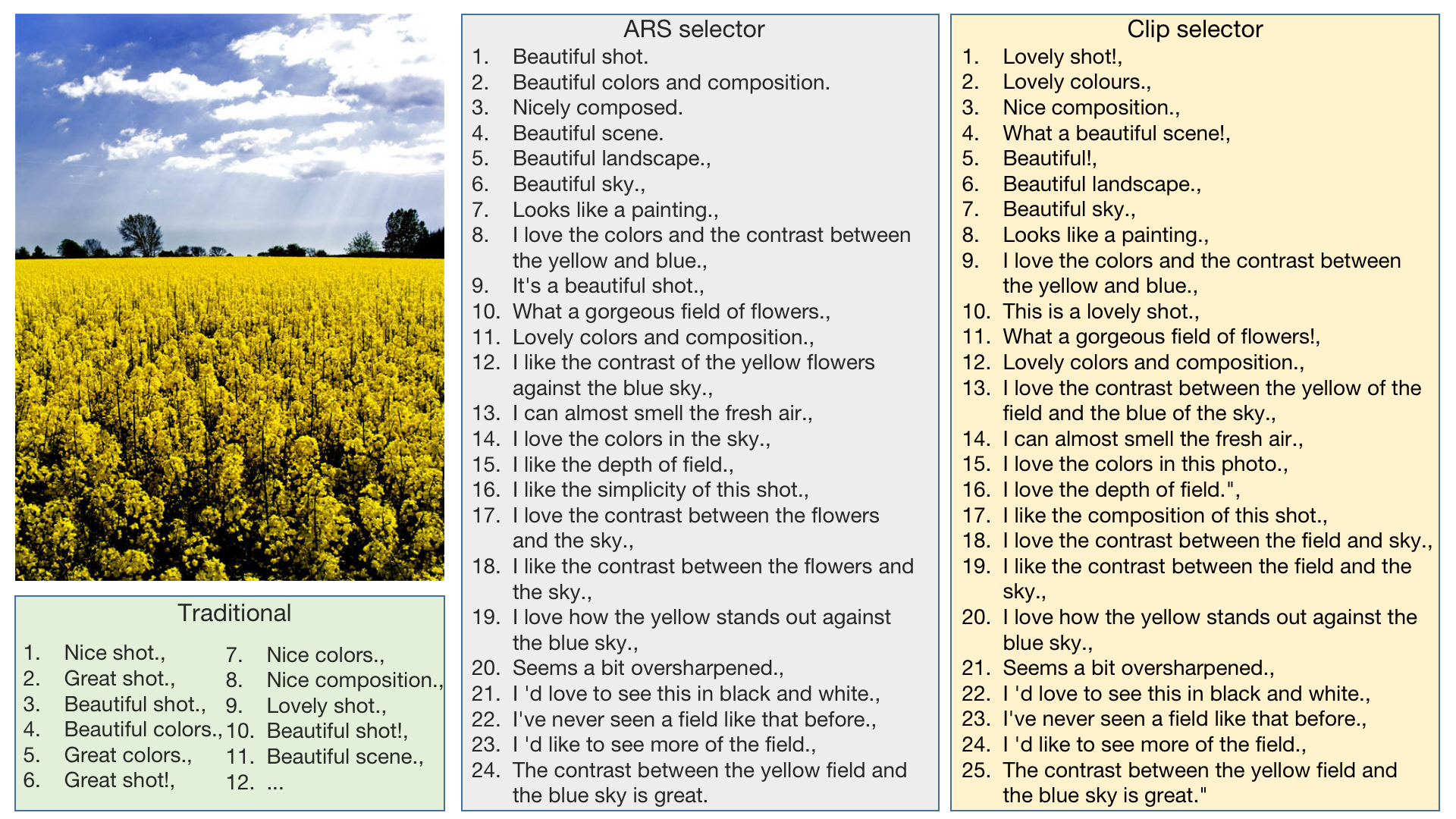} 
    \caption{Example captions. Traditional: The sentences are ranked based on the generator's confidence without using the DACS. ARS selector: \textit{ARS} is used to construct the DACS and pick the generated sentences according to \textit{ARS} values. CLIP selector: the fine tuned CLIP model is used to rank the generated sentences to construct the DACS and pick the sentences according to the CLIP matching scores. Note CLIP is only used to pick the best sentence in a group as one of the output sentences of DACS. It is seen that with DACS, the captions are more diverse. More example results are available in the Supplementary materials. }
	\label{fig-diversityresult15}
\end{figure*}

As described in the main method, with the introduction of \textit{ARS}, we can use \textit{ARS} based diverse aesthetic caption selector (DACS) to generate a diverse set of image captions rather than being restricted to output only one single sentence as in previous methods \cite{chang2017aesthetic}. Figures \ref{fig-diversityresult15} shows examples of aesthetic captions generated with \textit{ARS} based and CLIP based DACS. More examples are available in the Supplementary materials.

\begin{table}[th]
	\begin{center}
		\resizebox{\linewidth}{!}{
			\begin{tabular}{|c|c|c|c|c|}
				\hline
				Methods &  METETOR $\uparrow$  & ROUGE(L) $\uparrow$ & CIDER$\uparrow$ & SPICE$\uparrow$   \\ \hline
				Baseline & 0.1227 & 0.3543  & 0.0580 & 0.0172 \\ \hline
				ARIC (\ref{WeightedLossFunction}) & 0.1389 & 0.3610  & 0.0633 & 0.0353 \\\hline
		\end{tabular}}
		\caption{Image aesthetic captioning performances of our new model as compared with the baseline model. }
		\label{tab:infromation comment score improve caption improvement.}	
	\end{center}
\end{table}

%

\begin{table}[h]
	\begin{center}
		\resizebox{\linewidth}{!}{
			\begin{tabular}{|c|c|c|c|c|c|}
				\hline
				Method    & ACC$\uparrow$    & SRCC$\uparrow$   & PLCC$\uparrow$   & RMSE$\downarrow$    & MAE $\downarrow$  \\ \hline
				Image only & 0.8194  & 0.6755 & 0.6868 & 0.5356  & 0.4193 \\ \hline
				\multicolumn{6}{|c|}{AQA using Ground Truth Text} \\ \hline
				Text only & 0.8533 & 0.8274 & 0.851 & 0.3828 & 0.2960 \\ \hline
				Multi-modal(Image+Text) & 0.8407 &0.7529 & 0.7691 & 0.4650 & 0.3659 \\ \hline
				\multicolumn{6}{|c|}{AQA using Generated Captions} \\ \hline
				Text only Traditional & 0.7630 & 0.4335 & 0.4269 & 0.8921 & 0.7152 \\ \hline
				Text only DACS (ARS) & 0.7695 & 0.4711 & 0.4693 & 0.7881 & 0.6227 \\ \hline
				Text only DACS (CLIP) & 0.7720 & 0.476 & 0.4745 & 0.7645 & 0.6018 \\ \hline

			    Multi-modal Traditional & 0.8181 & 0.6818 & 0.6925 &0.5422 & 0.4232 \\ \hline
				Multi-modal DACS (ARS) &0.8195 & 0.6833 & 0.6934 & 0.5357 & 0.4192 \\ \hline
				Multi-modal DACS (CLIP) & 0.8203 & 0.6837 & 0.6940 & 0.5340 & 0.4179 \\ \hline

		\end{tabular}}
		\caption{Image AQA using generated captions. Traditional: no aesthetic relevance selection where all generated sentences are used. DACS(ARS): using sentences selected by the ARS based diverse aesthetic caption selector (DACS). DACS(CLIP): using sentences selected by the CLIP based diverse aesthetic caption selector (DACS).}
		\label{tab:generation caption aesthetic assessment v2}	
	\end{center}
\end{table}

\subsection{Image AQA based on Generated Captions}

In this experiment, we evaluate image AQA performances based on the generated image captions and results are shown in Table \ref{tab:generation caption aesthetic assessment v2}. We can observe that in text only AQA, the generated texts perform worse than the true captions, and also worse than image only AQA. It is also seen that using captions selected by the new diverse aesthetic caption selector (DACS) either based on ARS or CLIP performs better than using captions without selection. It is interesting to observe that in multi-modal AQA, including the generated texts has started to slightly exceed image only AQA, but is still quite far from those using ground truth. For example, using ground truth text, the ACC of multi-modal AQA is 0.8407 whilst that using generated captions is 0.8203. Would a future image captioning model be able to generate captions to close the gap? This would be a very interesting question and a goal for future research. 

\section{Concluding Remarks}
In this paper, we have attempted to study two closely related subjects of aesthetic visual computing, image aesthetic quality assessment (AQA) and image aesthetic captioning (IAC). We first introduce the concept of aesthetic relevance score (ARS) and use it to design the aesthetically relevant image captioning (ARIC) model through an ARS weighted loss function and an ARS based diverse aesthetic caption selector (DACS). We have presented extensive experimental results which demonstrate the soundness of the ARS concept and the effectiveness of the ARIC model. We have also contributed a large research database DPC2022 that contains images with both comments and aesthetic ratings.

\section{Acknowledgements}
This work was supported in part by the National Natural Science Foundation of China under grants U22B2035 and 62271323, in part by Guangdong Basic and Applied Basic Research Foundation under grant 2021A1515011584, and in part by the Shenzhen Research and Development Program under grants JCYJ20220531102408020 and JCYJ20200109105008228).

\bibliography{aaai23.bib}

\clearpage

\appendix

\begin{figure*}[h]
	\centering
	\includegraphics[width=1.0\textwidth]{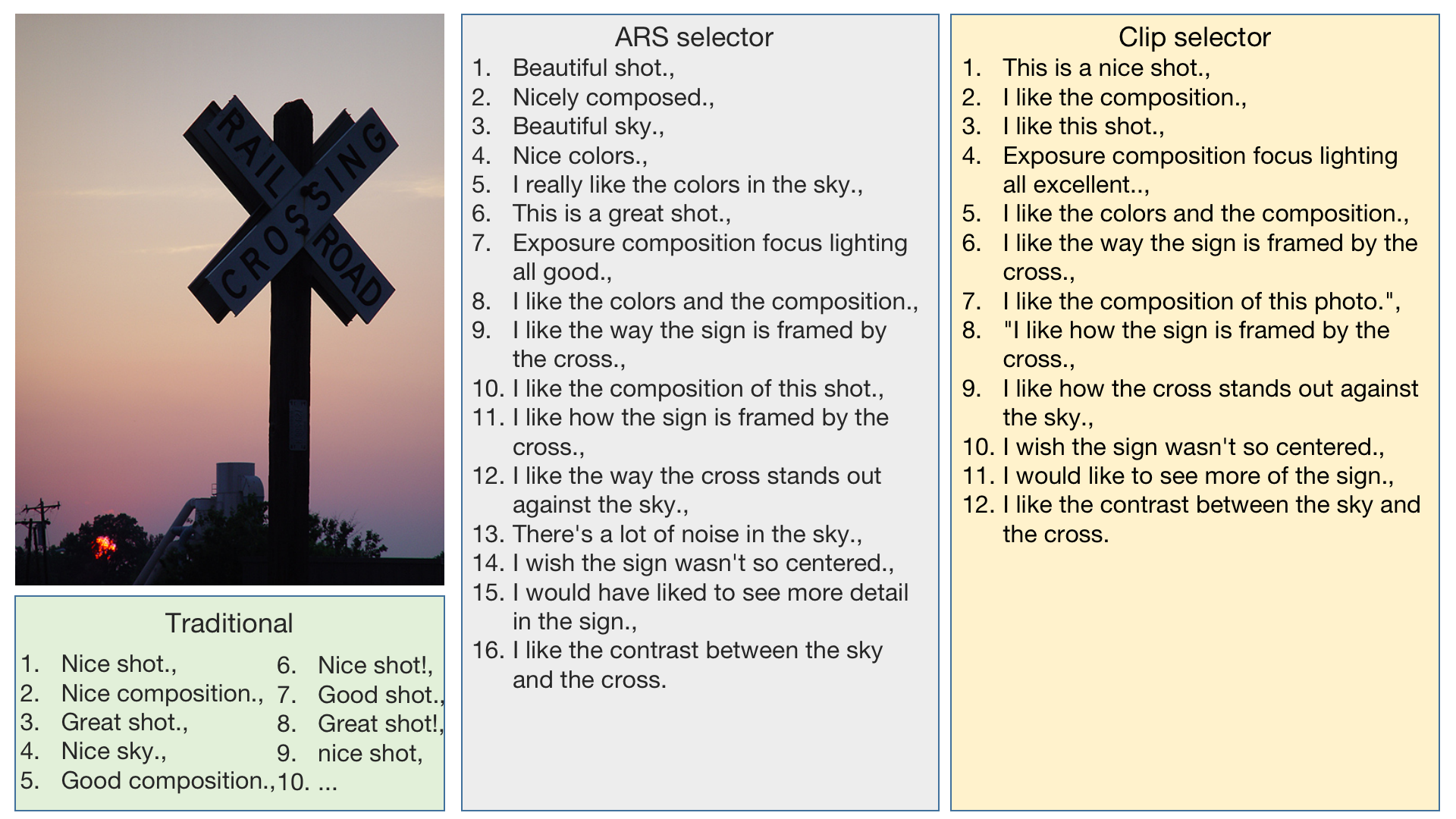}  
	\caption{More examples}
	\label{fig-diversityresult1}
\end{figure*}

\begin{figure*}[h]
	\centering
	\includegraphics[width=1.0\textwidth]{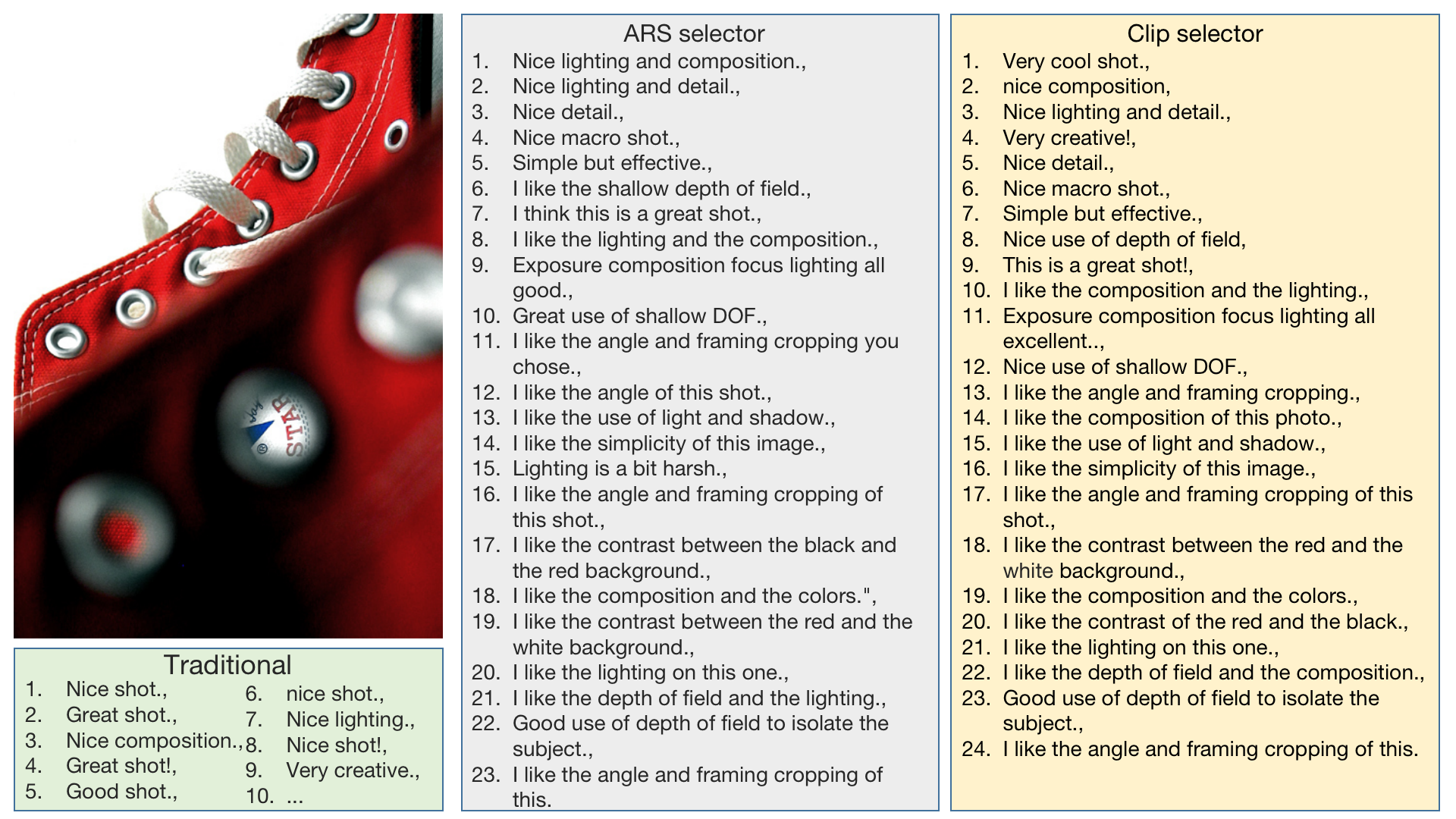}  
	\caption{More examples}
	\label{fig-diversityresult2}
\end{figure*}

\begin{figure*}[tph]
	\centering
	\includegraphics[width=1.0\textwidth]{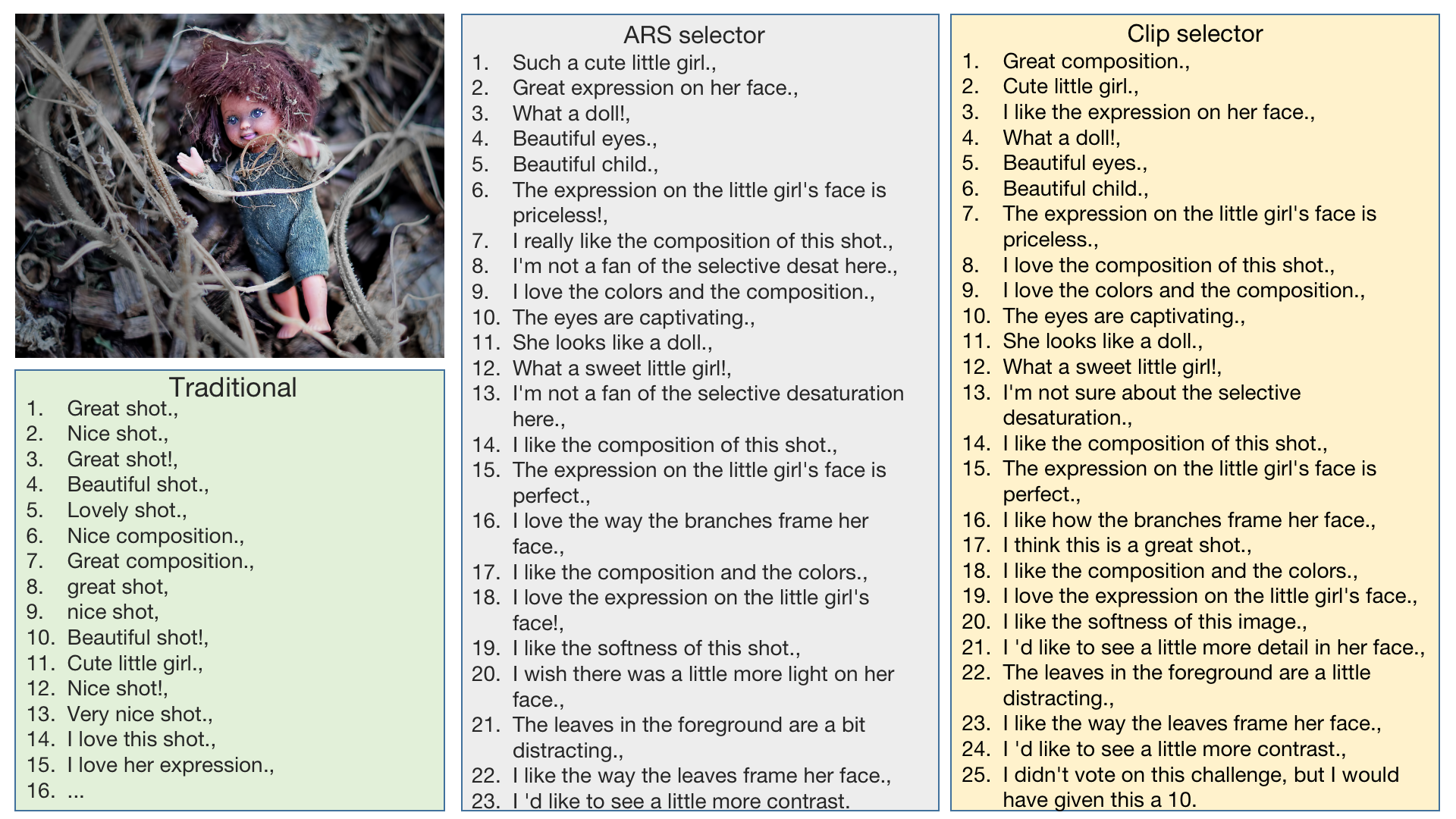}  
	\caption{More examples}
	\label{fig-diversityresult3}
\end{figure*}

\begin{figure*}[tph]
	\centering
	\includegraphics[width=1.0\textwidth]{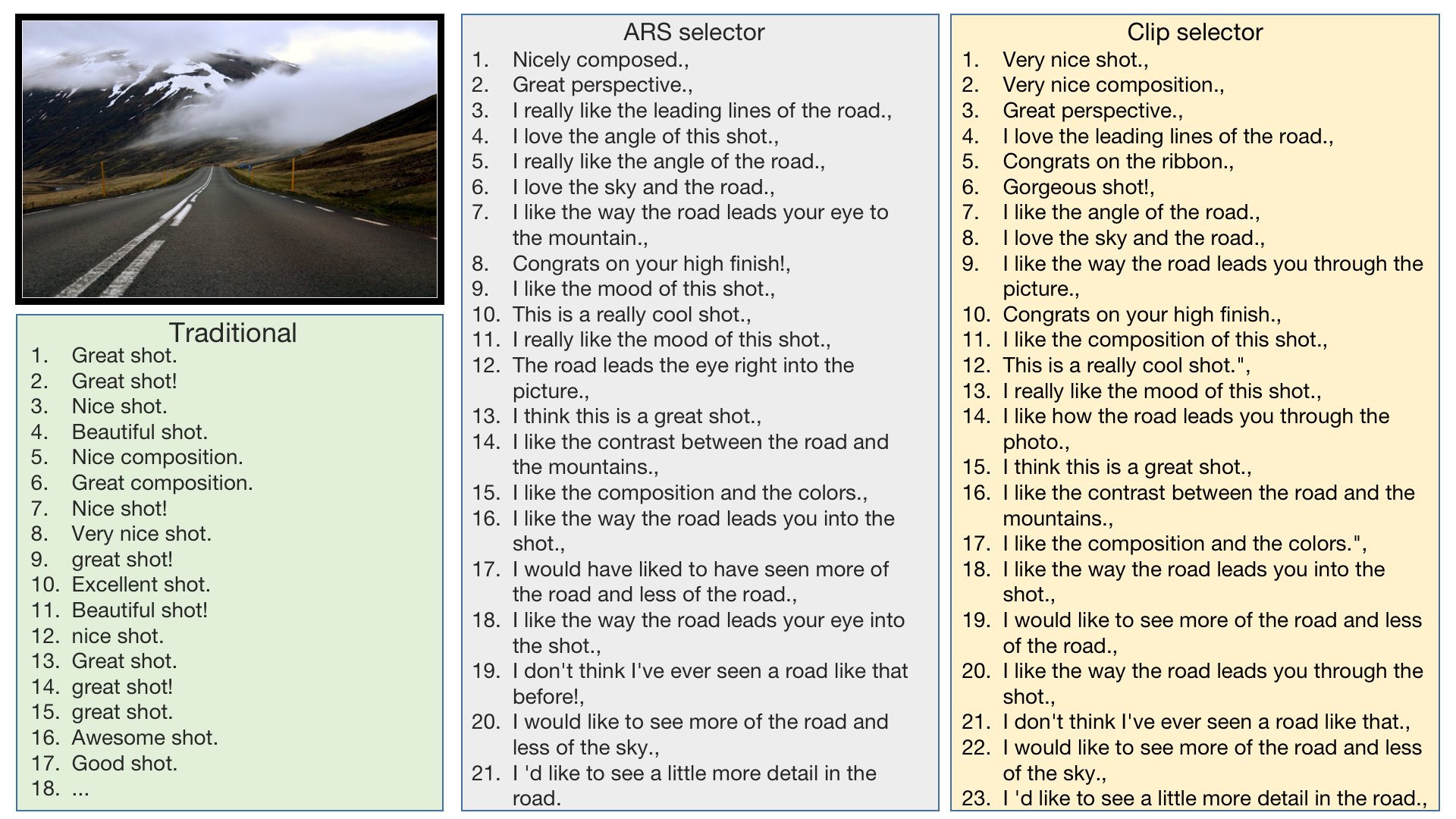}  
	\caption{More examples}
	\label{fig-diversityresult4}
\end{figure*}

\begin{figure*}[tph]
	\centering
	\includegraphics[width=1.0\textwidth]{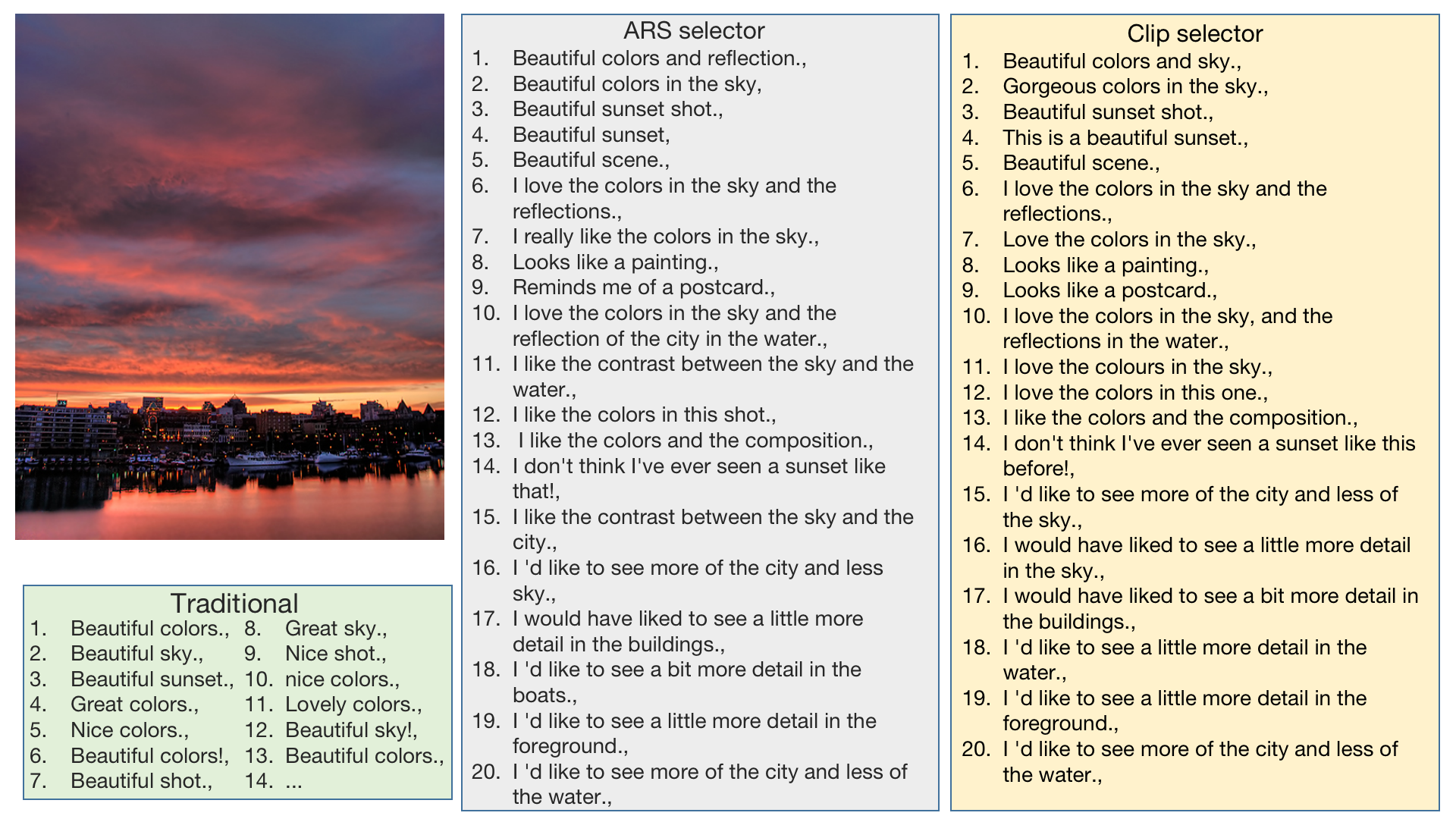}  
	\caption{More examples}
	\label{fig-diversityresult5}
\end{figure*}

\begin{figure*}[tph]
	\centering
	\includegraphics[width=1.0\textwidth]{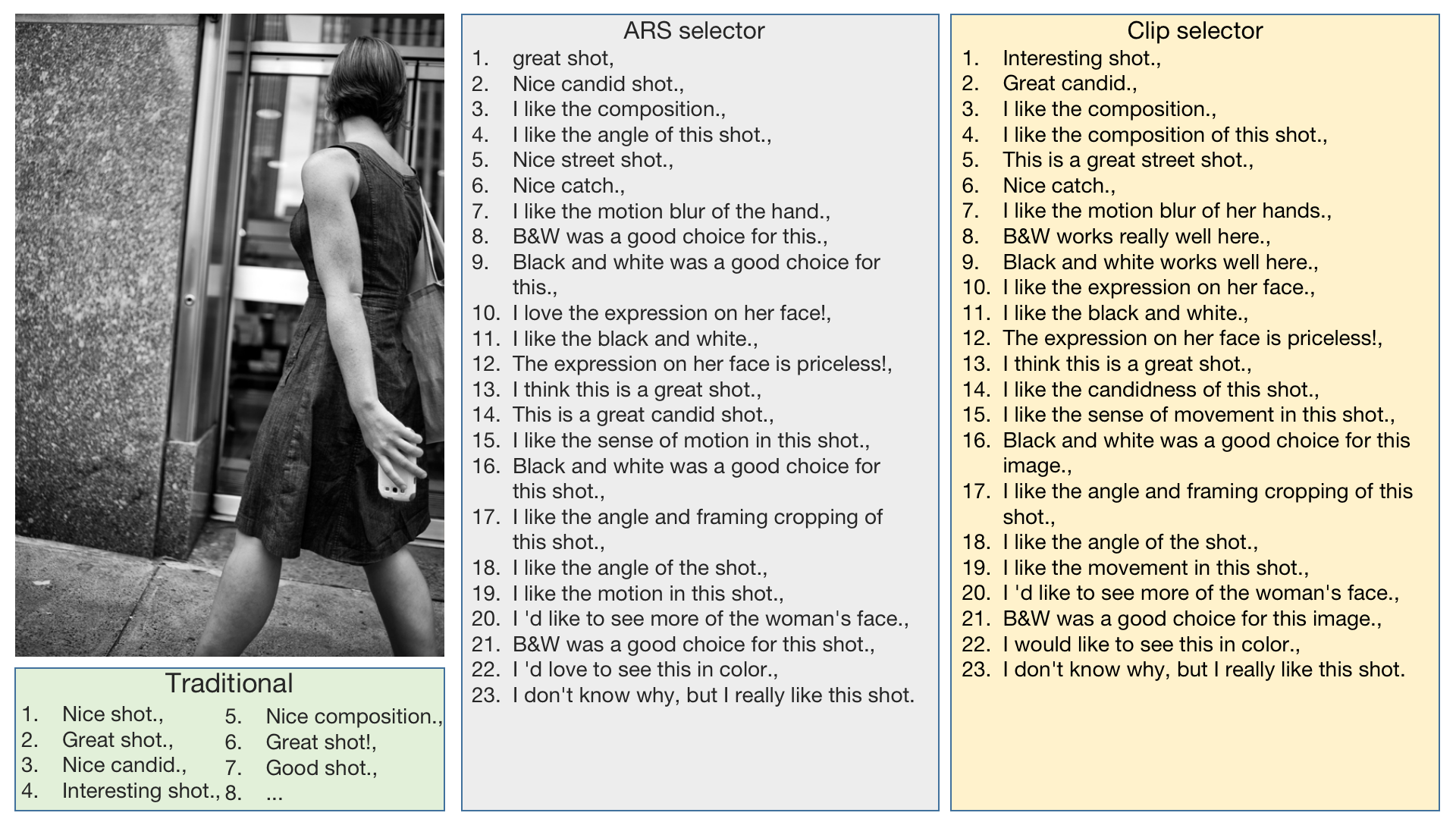}  
	\caption{More examples}
	\label{fig-diversityresult6}
\end{figure*}

\section{Appendix I} 


\label{section:datalabel_supplementary_material}
The \textit{ARS} of a sentence $t$ is defined as: 
 \begin{equation}
 \label{ARD_definition_supplementary_material}
 ARS (t) = A(t)+L(t)+ O(t) + S(t)  + T_{fidf}(t)
 \end{equation}
 
where $A(t)$ is related to aesthetic words, $L(t)$ is related to the length of $t$, $O(t)$ is related to object words, $S(t)$ is the sentiment score of $t$, and $T_{fidf}(t)$ is related to term frequency–inverse document frequency.  

We first counted the nouns, adverbs, adjectives, and verbs of all comments, sorted by word frequency from high to low. We manually selected 1022 most frequently appeared image aesthetics related words, the full list, $\{AW_{list}\}$, can be found in \textbf{Appendix II}. We also manually selected 2146 words related to objects, the full list of these words, $\{OW_{list}\}$, can be found in \textbf{Appendix III}.

To obtain the $A(t)$, we simply count how many of these 1022 words have appeared in $t$

\begin{equation}
\label{A(t)_definition}
\begin{aligned}
A(t) = \sum_{w=1}^{w=|t|} \delta_a (t_w),  
\end{aligned}
\end{equation}

where $t_w$ is the $w^{th}$ word in $t$, $|t|$ is the length of $t$, $\delta_a (t_w)=1$ if $t_w\in AW_{list}$ and  $\delta_a (t_w)=0$ otherwise.   

$L(t)$ is calculated as 
 \begin{equation}
\label{L(t)_definition}
\begin{aligned}
L(t) = \frac{\beta(|t|, m_{|t|}, \sigma_{|t|})-\beta(|t|_{min}, m_{|t|}, \sigma_{|t|})} {\beta(|t|_{max}, m_{|t|}, \sigma_{|t|})-\beta(|t|_{min}, m_{|t|}, \sigma_{|t|})}
\end{aligned}
\end{equation}

where $|t|$ is the length (number of words) of $t$, $m_{|t|}$ is the average length of sentences, $\sigma_{|t|}$ is the variance of the length of the sentences, $|t|_{max}$ is the length of the longest sentence and $|t|_{min}$ is the length of the shortest sentence in the  dataset, and $\beta$ is a sigmoid function defined in (\ref{sigmoid}) 

\begin{equation}
\label{sigmoid}
\begin{aligned}
\beta(x, m, \sigma) = \frac{1}{1+ e^{-(\frac{x-m}{\sigma})}}
\end{aligned}
\end{equation}

$O(t)$ is calculated as 
\begin{equation}
\label{O(t)_definition}
O(t) = \sum_{w=1}^{w=|t|} \delta _o(t_w)
\end{equation}
where $t_w$ is the $w^{th}$ word in $t$, $|t|$ is the length of $t$, $\delta_o (t_w) = 1$  if $t_w\in OW_{list}$, and $\delta_o (t_w) = 0$ otherwise.

$S(t)$ is the sentiment score of $t$. We used the $BerTweet$ model and the toolkit from\cite{perez2021pysentimiento} to calculate both the positive and negative sentiments of $t$ and obtain $S(t)$ as 
\begin{equation}
\label{S(t)_definition}
S(t) = 	\frac{P_s(t) + N_s(t)}{2}
\end{equation}
where $P_s(t)$ is the predicted positive sentiment and $N_s(t)$ is the predicted negative sentiment of $t$, and the values of both $P_s(t)$ and $N_s(t)$ are in $[0, 1]$.   

$T_{fidf}(t)$ is related to term frequency-inverse document frequency (\textit{tf-itf}). For a give term \textit{tm}, we first calculate its  \textit{tf-itf} value $\tau(tm)$

\begin{equation}
\begin{aligned}
\tau(tm)=\frac{n_{tm}}{N_{tm}}*\left(log\left(\frac{1+N}{1+I_{tm}}\right)+1\right)
\end{aligned}
\end{equation}
where $n_{tm}$ is the number of times the term $tm$ has appeared in the comment of a given image, $N_{tm}$ is the total number of terms in the comment of the image, $N$ is the total number of images in the dataset, and $I_{tm}$ is the number of images with comments containing the term $tm$. We then normalise $\tau(tm)$ according to

\begin{equation}
\label{Normalizing tfidf}
\begin{aligned}
\tau_n(tm) =\frac{\beta(\tau(tm), m_{\tau(tm)}, \sigma_{\tau(tm)})-\beta_{min}(\tau(tm))} {\beta_{max}(\tau(tm))-\beta_{min}(\tau(tm))}
\end{aligned}
\end{equation}

where $\beta_{min}(\tau(tm))=\beta(\tau_{min}(tm), m_{\tau(tm)}, \sigma_{\tau(tm)})$, $\beta_{max}(\tau(tm)) = \beta(\tau_{max}(tm), m_{\tau(tm)}, \sigma_{\tau(tm)})$, $\tau_{min}(tm)$ is the smallest $\tau(tm)$, $\tau_{max}(tm)$ is the largest $\tau(tm)$, $m_{\tau(tm)}$ is the mean value and $\sigma_{\tau(tm)}$ is the variance of $\tau(tm)$ of the dataset. 

Finally, we compute the $T_{fidf}(t)$ in equation (\ref{ARD_definition}) as
\begin{equation}
\label{tfidf_score}
\begin{aligned}
T_{idf}(t)=  \sum_{i=1}^{i=|t|} \tau_n(tm(i))
\end{aligned}	
\end{equation}
where $tm(i)$ is the $i^{th}$ term, $|t|$ is the length of $t$. 

\begin{table*}[]
	\begin{center}
		\resizebox{\linewidth}{!}{
			\begin{tabular}{|c|c|c|c|c|c|c|c|}
				\hline
				\multicolumn{8}{|c|}{\textbf{Appendix II: Aesthetic word table, 1/2}} \\  \hline
				shot & color & composition & light & focus & lighting & background & subject \\ \hline	
right & detail & shoot & contrast & crop & colour & tone & white \\ \hline	
blue & dark & line & texture & angle & shadow & sharp & frame \\ \hline	
black & high & left & reflection & soft & point & post & red \\ \hline	
perspective & overall & distract & leave & strong & bright & creative & scene \\ \hline	
exposure & low & camera & simple & green & blur & view & space \\ \hline	
away & concept & border & catch & foreground & depth & long & small \\ \hline	
abstract & spot & center & yellow & motion & area & tight & draw \\ \hline	
lack & shape & moment & lead & neat & attention & miss & dof \\ \hline	
balance & quality & field & mood & sharpness & element & distracting & edge \\ \hline	
highlight & theme & level & compose & end & stop & corner & art \\ \hline	
rest & flat & framing & tilt & pattern & bokeh & lens & cut \\ \hline	
large & move & unique & blurry & horizon & rule & noise & silhouette \\ \hline	
warm & grain & shallow & movement & middle & curve & size & pull \\ \hline	
orange & cropping & saturation & sharpen & expose & fill & emotion & odd \\ \hline	
couple & grainy & range & cover & style & sepia & similar & position \\ \hline	
focal & shutter & softness & simplicity & half & branch & suggest & pink \\ \hline	
imagine & hot & filter & overexposed & match & body & form & brown \\ \hline	
classic & diagonal & distraction & remove & purple & square & impressive & reflect \\ \hline	
colorful & stick & smooth & blend & combination & location & shade & mirror \\ \hline	
layer & vertical & impression & snapshot & centered & desat & darkness & suit \\ \hline	
overlay & aspect & single & grey & tonal & flow & hue & connection \\ \hline	
base & entire & blurred & higher & symmetry & part & gray & distance \\ \hline	
coloring & saturate & backdrop & scale & noisy & surface & balanced & emotive \\ \hline	
horizontal & cross & structure & golden & rotate & atmosphere & direction & context \\ \hline	
outside & fade & btw & aperture & gold & pure & pixel & thread \\ \hline	
wire & centre & bunch & fuzzy & film & profile & focused & toning \\ \hline	
mask & environment & desaturation & mystery & inside & portion & side & mysterious \\ \hline	
crowd & mix & b\&w & duotone & order & addition & colored & emotional \\ \hline	
iso & underexposed & central & underrated & bg & architecture & ray & oversharpened \\ \hline	
pastel & detailed & surround & group & mess & combine & apart & aside \\ \hline	
sport & narrow & cluttered & tint & band & spirit & desaturate & distortion \\ \hline	
darken & artifact & compression & dimension & compostion & closeup & upside & major \\ \hline	
compositionally & positioning & contrasty & repeat & fork & tall & lay & triangle \\ \hline	
reference & juxtaposition & condition & surrounding & shooting & ripple & weak & belong \\ \hline	
lighten & placing & connect & ghost & warmth & pair & row & viewpoint \\ \hline	
sight & lightning & stripe & backlighting & clue & top & minimalism & colourful \\ \hline	
intensity & continue & hdr & soften & brighten & orientation & layout & lower \\ \hline	
tonality & outline & rotation & obscure & sensor & lense & oversaturated & height \\ \hline	
blind & resolution & artificial & fisheye & palette & parallel & geometry & cropped \\ \hline	
crooked & silver & partial & combo & overprocessed & makeup & offset & streak \\ \hline	
patch & wrinkle & resize & concentrate & minimalist & symmetrical & saturated & backround \\ \hline	
length & raw & monochrome & coloration & misty & criterion & tightly & uneven \\ \hline	
up & hazy & rhythm & pale & oversharpene & relationship & panning & recommend \\ \hline	
graininess & constructive & rear & textured & forground & geometric & overexposure & blank \\ \hline	
unbalanced & posture & unsharp & aesthetic & rim & backlit & channel & interaction \\ \hline	
uniform & smoothness & proportion & sentiment & foggy & cloudy & isolation & ratio \\ \hline	
speck & cell & colouring & lensbaby & overcast & backlighte & blob & zone \\ \hline	
blurring & regular & polarizer & circular & drab & overexpose & bw & back \\ \hline	
minimalistic & coloured & compositon & backlight & overhead & abstraction & partially & composite \\ \hline	
curved & oversaturate & greenish & reshoot & sharply & slant & spotlight & fore \\ \hline	
pix & yellowish & compo & calibrate & cartoon & cone & panoramic & centred \\ \hline	
vertically & haloing & align & slope & spiral & interior & monochromatic & vibe \\ \hline	
exotic & cube & impressionistic & arrow & sharpening & horizontally & photojournalism & backwards \\ \hline	
sized & bluish & pixelate & refraction & monotone & reddish & indoor & greenery \\ \hline	
watercolor & ink & cherry & magenta & stance & midtone & blending & mixed \\ \hline	
photoshoppe & cyan & toned & nostalgia & whiter & faded & panorama & tinge \\ \hline	
underexpose & shot- & overlap & gloomy & focusse & outdoor & runner & container \\ \hline	
vista & transparency & tune & lightness & cam & symetry & rectangle & width \\ \hline	
shading & aperature & polish & dome & brightly & aesthetically & trim & tuning \\ \hline	
pixelated & silhoutte & shadowy & angled & mixture & grid & esque & blurriness \\ \hline	
overprocesse & noir & gradation & grayscale & skew & converge & slanted & tan \\ \hline	
sphere & blackness & pixelation & compress & trend & straightforward & interplay & unfocused \\ \hline	
upwards & emotionally & shadowing & histogram & translucent & comic & bent & impressionist \\ \hline	
scatter & shut & gaussian & futuristic & arc & illumination & temperature & radiate \\ \hline	
opacity & metaphor & abstractness & alignment & underside & softly & linear & pinkish \\ \hline	
vague & textural & void & resizing & ultra & bias & realism & vertigo \\ \hline
		\end{tabular}}
		\caption{Aesthetic word table}
		\label{tab:related to aesthetic words table supplementary material 3}	
	\end{center}
\end{table*}

\begin{table*}[]
	\begin{center}
		\resizebox{\linewidth}{!}{
			\begin{tabular}{|c|c|c|c|c|c|c|c|}
				\hline
				\multicolumn{8}{|c|}{\textbf{Appendix II: Aesthetic word table, 2/2}} \\  \hline
fringe & sharper & silouette & mesh & boundary & barrier & torso & simplify \\ \hline	
blurr & artistry & postprocessing & sword & impressionism & span & mapping & romance \\ \hline	
backside & silvery & intersect & polarizing & copper & surrealism & juxtapose & diagonally \\ \hline	
luminosity & backgound & triangular & contour & solitary & greyscale & whiteness & similarity \\ \hline	
symetrical & violet & scope & blueish & amber & compostition & twilight & olive \\ \hline
ambiguous & hanger & vastness & loop & sentimental & indistinct & sillhouette & contrasted \\ \hline	
datum & fuzz & dichotomy & curvy & surrealistic & brownish & rectangular & dizzying \\ \hline	
blured & shelter & interfere & wavy & shimmer & symmetric & sillouette & shooter \\ \hline	
artefact & bleach & smoothing & orb & b+w & blotchy & colortone & gamma \\ \hline	
curvature & contrary & blemish & prespective & refection & viewfinder & inclined & oval \\ \hline	
cartoonish & restriction & fuzziness & chromatic & fringing & underexposure & lighing & focusing \\ \hline	
balancing & bronze & flatness & beige & resemblance & borderline & aqua & contrasting \\ \hline	
nighttime & splotch & shaded & perpendicular & radial & masking & radius & verticle \\ \hline	
ambiance & bacground & uv & ambience & calibration & swirly & excess & dazzle \\ \hline	
hightlight & backgroud & softbox & coil & unbalance & oversharp & detailing & bluer \\ \hline	
oversaturation & axis & refract & softening & grayish & purplish & granular & verge \\ \hline	
tonemapping & distracted & shutterspeed & tilting & midst & understatement & optic & compressed \\ \hline	
perpective & diffusion & treeline & superimpose & shaped & brightening & tendency & askew \\ \hline	
lighitng & nuance & specular & refelction & roundness & peacefulness & nothingness & ramp \\ \hline	
affection & rigid & similarly & imbalance & coordinate & asymmetrical & reflected & stylize \\ \hline	
artifacte & grandeur & skewsme & roughness & greyish & blacker & highlighting & nd \\ \hline	
mosaic & camo & silohuette & bound & dof. & oversharpening & buff & rad \\ \hline	
colorfull & compozition & fleck & pallette & compsition & silouhette & concentrated & overshadow \\ \hline	
asymmetry & variance & geometrical & spacing & mystique & winding & shot-- & autofocus \\ \hline	
bluey & posed & lightsource & particle & projection & tonemappe & leftmost & compisition \\ \hline	
opaque & autumnal & grainyness & incline & deatil & oblique & haziness & rightmost \\ \hline	
prism & luminance & reflexion & blueness & criss & colorless & apeture & emerald \\ \hline	
apparition & clearness & watercolour & relfection & coordination & redish & balck & colorization \\ \hline	
vacant & artifical & straighter & hemisphere & ligthing & colouration & zooming & gimmick \\ \hline	
evergreen & backgroung & uniformly & orangy & beneath & uniformity & bleakness & superfluous \\ \hline	
reflex & affective & framework & radioactive & mirroring & translucence & epitome & patina \\ \hline	
teal & cartoony & tinting & shear & silhouetting & foucs & imaging & polarize \\ \hline	
veiw & polariser & texturing & splotchy & midground & crunch & difuse & silhoette \\ \hline	
sillouhette & distractingly & foregound & downsize & croping & translucency & postprocesse & peachy \\ \hline	
f2.8 & 800px & visibility & backstory & constitute & storyline & composistion & colors \\ \hline	
ligting & boke & sephia & ballance & arched & coposition & overprocessing & spherical \\ \hline	
uncropped & highlite & wideangle & hilight & composition- & pixellate & colorize & darkish \\ \hline	
spotted & deform & camoflauge & yellowy & orangish & highlighted & harmonize & overlit \\ \hline	
colors- & whitebalance & index & oversized & sidelighte & darkest & misplace & speckled \\ \hline	
sharpne & backgorund & tilted & brokeh & backgroun & themed & reshot & cleanliness \\ \hline	
sharpend & denoise & undertone & diffraction & sihouette & blurri & pixle & pixellation \\ \hline	
persepctive & texure & infocus & uncalibrated & focuse & marginal & phantom & composing \\ \hline	
noice & comosition & inclination & caliber & matt & higlight & greyness & projector \\ \hline	
outlook & megapixel & detial & composotion & horison & polarising & unevenly & clairity \\ \hline	
squarish & collor & pixilated & topple & luster & dun & shininess & subect \\ \hline	
mauve & dslr & fuzzie & sharpeness & gradiation & discoloration & refelection & colorwise \\ \hline	
ccd & chiaroscuro & oversharpen & gradiant & colorized & apperture & sharpener & abberation \\ \hline	
shot! & exposer & indigo & exposure(2 & bluriness & blurryness & colore & maroon \\ \hline	
colourfull & horizion & skiprow & crimson & harmonic & leveling & focus- & glob \\ \hline	
despeckle & sigma & unprocessed & feathering & focussing & refinement & resonance & greeny \\ \hline	
color- & coppery & quad & blacklight & fidelity & hyperfocal & ghosting & composiiton \\ \hline	
telegraph & symetric & persective & dizziness & symmetrically & overlaid & straightness & delineation \\ \hline	
beegee & clipping & pallete & compositing & distrace & siloette & reprocess & colrs \\ \hline	
quadtone & underlit & antithesis & stripey & streche & taper & colorsaturation & grayness \\ \hline	
rightside & grainey & hubcap & unsaturated & speedlight & fuschia & unsharpness & coulour \\ \hline	
distrating & geometrically & multicolore & shawdow & pespective & clog & undersaturated & shodow \\ \hline	
brink & overexposured & brighness & bakground & tolerant & colourisation & diagnal & coor \\ \hline	
middleground & prospect & backgrond & orginality & whitespace & colro & b\&w. & ringlight \\ \hline	
recroppe & tolerance & balence & granule & perspecitve & disctracte & latex & evenness \\ \hline	
defocuse & compositin & light+idea+execution & emo & siloutte & conrast & softfocus & fuzze \\ \hline	
deatail & unevenness & croped & compesition & compositiion &   &   &   \\ \hline
		\end{tabular}}
		\caption{Aesthetic word table}
		\label{tab:related to aesthetic words table supplementary material 1}	
	\end{center}
\end{table*}



\begin{table*}[]
	\begin{center}
		\resizebox{\linewidth}{!}{
			\begin{tabular}{|c|c|c|c|c|c|c|c|}
				\hline
				\multicolumn{8}{|c|}{\textbf{Appendix III: Object word table, 1/5}} \\  \hline
				eye & sky & face & ribbon & water & tree & flower & expression \\ \hline	
people & hand & bird & head & story & sense & glass & guy \\ \hline	
man & dog & hair & girl & wall & cat & smile & sun \\ \hline	
window & car & rock & crisp & kid & landscape & fall & child \\ \hline	
skin & drop & snow & hang & fan & house & book & grass \\ \hline	
bridge & ground & painting & horse & sunset & street & road & fence \\ \hline	
boat & baby & woman & animal & leg & ball & fly & leaf \\ \hline	
boy & shoe & wing & arm & paint & room & bottle & home \\ \hline	
mountain & box & wood & star & door & city & moon & plant \\ \hline	
smoke & feather & nose & hat & finger & bubble & paper & ice \\ \hline	
fire & family & petal & fish & shirt & table & glare & text \\ \hline	
card & human & sand & duck & chair & lady & brightness & total \\ \hline	
zoom & rain & beach & wave & apple & egg & ring & dress \\ \hline	
train & circle & fog & food & land & stone & wind & mouth \\ \hline	
floor & ear & bike & halo & flame & butterfly & plane & bee \\ \hline	
wheel & cow & portfolio & round & tie & magazine & spider & sea \\ \hline	
metal & dust & pencil & pet & kitty & cup & phone & postcard \\ \hline	
tower & flag & droplet & statue & weather & neck & hill & toy \\ \hline	
shoulder & lamp & energy & candle & lip & rose & mist & board \\ \hline	
crack & creature & fur & bar & tail & summer & pan & plastic \\ \hline	
sunlight & brick & fruit & trail & church & coffee & drink & rainbow \\ \hline	
boot & brush & country & earth & candy & plate & hole & roof \\ \hline	
money & music & tack & scenery & material & block & umbrella & tooth \\ \hline	
frog & album & urban & milk & bowl & balloon & bench & shell \\ \hline	
tear & insect & spring & truck & doll & flight & town & barn \\ \hline	
sucker & waterfall & chocolate & artist & garden & bag & dirt & oof \\ \hline	
lighthouse & river & dance & beer & guitar & clock & wine & sunrise \\ \hline	
lake & vase & button & build & cloth & chain & storm & tongue \\ \hline	
clothe & bulb & liquid & alien & ocean & ship & rope & cap \\ \hline	
font & stair & park & sheet & bloom & cheese & flaw & wedding \\ \hline	
fault & bush & beak & arch & soul & rail & gun & catchlight \\ \hline	
crystal & pin & shop & trunk & machine & spoon & knife & seed \\ \hline	
sculpture & puzzle & blood & bed & angel & chin & snake & fabric \\ \hline	
oil & coat & deer & foliage & woody & ribbone & squirrel & cheek \\ \hline	
pepper & toe & puppy & wildlife & cake & seat & ceiling & tv \\ \hline	
bud & shoehorn & pool & skyline & curtain & forest & haze & pot \\ \hline	
mouse & cream & butt & hammer & coin & swan & clothing & stream \\ \hline	
sheep & pier & swirl & outfit & kitchen & jacket & banana & autumn \\ \hline	
canvas & steam & yard & lock & vehicle & palm & lemon & pigeon \\ \hline	
planet & salt & chicken & critter & stack & tulip & pen & dish \\ \hline	
gem & newspaper & cookie & tape & forehead & ant & strawberry & gull \\ \hline	
clip & bread & tunnel & needle & artwork & pipe & billboard & berry \\ \hline	
knee & fountain & dock & marble & garbage & tomato & cityscape & sugar \\ \hline	
chest & bow & spark & bean & horn & sock & cigarette & elephant \\ \hline	
straw & screw & pile & glove & mushroom & column & basket & owl \\ \hline	
eagle & pumpkin & bunny & pup & cd & net & blanket & anchor \\ \hline	
iron & carpet & jewelry & bride & gate & lion & chip & ornament \\ \hline	
fishing & dragonfly & kitten & nut & costume & castle & hood & snail \\ \hline	
grape & twig & blossom & crisper & stamp & desert & graffiti & beast \\ \hline
railing & windmill & corn & sunglass & chrome & turtle & monster & station \\ \hline	
		\end{tabular}}
		\caption{Object word table}
		\label{tab:related to object words table supplementary material 4}	
	\end{center}
\end{table*}

\begin{table*}[]
	\begin{center}
		\resizebox{\linewidth}{!}{
			\begin{tabular}{|c|c|c|c|c|c|c|c|}
				\hline
				\multicolumn{8}{|c|}{\textbf{Appendix III: Object word table, 2/5}} \\  \hline
root & toilet & tube & jean & razor & breast & belly & pea \\ \hline	
farm & chess & lily & mug & vein & pond & tank & sink \\ \hline	
seagull & ducky & frost & diamond & bead & antique & skull & lizard \\ \hline
meter & sail & pig & museum & steel & nest & towel & golf \\ \hline	
peg & king & tattoo & jar & stalk & scarf & sidewalk & desk \\ \hline	
underwater & bone & island & bell & pillar & dancer & cliff & gas \\ \hline	
bank & bucket & necklace & crayon & pie & mud & dandelion & crane \\ \hline	
pear & cable & lipstick & rubber & cage & goose & surfer & puddle \\ \hline	
dragon & breeze & pod & stellar & collar & piano & pump & lace \\ \hline	
garage & baseball & shower & eyelash & craft & eyebrow & bicycle & leap \\ \hline	
jewel & rabbit & tractor & peel & vine & barrel & grog & ladder \\ \hline	
monument & sunshine & daisy & lunch & couch & moth & pine & headlight \\ \hline
pavement & iris & helmet & chick & dice & recipe & bat & doorway \\ \hline	
intersection & pearl & footprint & shelf & hummingbird & surf & meal & police \\ \hline	
cave & geese & airplane & pelican & bathroom & globe & bomb & strand \\ \hline	
shoreline & moss & plug & sunflower & guard & cowboy & skirt & daylight \\ \hline	
hook & strap & cord & sandy & circumstance & bleed & freckle & doggie \\ \hline	
battery & pupil & sleeve & sweater & bolt & telephone & alley & lime \\ \hline	
pane & lawn & scissor & butter & coast & peanut & decoration & stump \\ \hline	
flashlight & showcase & heel & bath & heron & kite & worm & walkway \\ \hline	
nuts & popcorn & deck & blury & pill & onion & shed & cart \\ \hline	
meat & pinhole & potato & ladybug & pocket & fisherman & bull & van \\ \hline	
earring & foam & dawn & skewed & silk & hay & crown & banding \\ \hline	
dusk & belt & veil & juice & vegetable & backyard & fluid & seal \\ \hline	
carving & pillow & mount & scape & leather & glue & grill & drug \\ \hline	
furniture & aquarium & eyeball & soldier & dune & universe & rat & pad \\ \hline	
wallpaper & smoking & sketch & cop & realm & windshield & mag & soup \\ \hline	
peacock & feast & velvet & mast & hose & reed & muscle & lash \\ \hline	
wrist & mannequin & flavor & cardboard & soap & wolf & zebra & sweat \\ \hline	
pollen & valley & village & orchid & poppy & fridge & hotel & jaw \\ \hline	
painter & powerline & pasta & violin & flood & doggy & football & wax \\ \hline	
figurine & package & raindrop & glint & vegetation & posing & flesh & resonate \\ \hline	
tent & rug & mat & fern & biker & tee & staircase & buck \\ \hline	
lid & junk & ridge & feeder & motorcycle & worker & dessert & concert \\ \hline	
grater & gesture & cigar & grasshopper & spike & doctor & rack & dancing \\ \hline	
teddy & mane & crab & pooch & loom & knob & claw & tombstone \\ \hline	
veggie & beetle & bullet & crow & giraffe & flora & actor & fox \\ \hline	
coke & chalk & drum & porch & hawk & cemetery & mural & robot \\ \hline	
pony & tin & bum & bra & shark & playground & outdoors & dam \\ \hline	
cabin & carrot & caterpillar & pedestrian & lavender & nap & cork & gum \\ \hline	
turkey & pitch & headstone & cyclist & bracelet & factory & spinning & tub \\ \hline	
rod & jelly & snowflake & lane & sailboat & tide & cement & toast \\ \hline	
wasp & rooster & railroad & wagon & gel & chimney & garlic & tissue \\ \hline	
canoe & kiwi & graveyard & peach & grit & canyon & lightbulb & blonde \\ \hline	
bin & port & sandwich & library & hall & cracker & rice & prison \\ \hline	
lantern & pineapple & princess & lava & downtown & pedal & dairy & sherpet \\ \hline	
wrench & soccer & bay & waterdrop & cathedral & pizza & charcoal & desktop \\ \hline	
radio & parrot & apartment & bouquet & cotton & icicle & camel & antler \\ \hline	
toad & turquoise & stamen & egret & tennis & memorial & helicopter & icing \\ \hline	
aurora & hydrant & tray & snack & hospital & spine & shrub & balcony \\ \hline	
nipple & spice & pilot & moonlight & topaz & salad & creek & temple \\ \hline	
midnight & aura & mall & diet & bamboo & bedroom & portait & airport \\ \hline	
whip & thorn & fauna & laser & soil & countryside & jam & pebble \\ \hline	
beverage & flake & walker & powder & lightening & mantis & aircraft & ham \\ \hline	

		\end{tabular}}
		\caption{Object word table}
		\label{tab:related to object words table supplementary matrial 2}	
	\end{center}
\end{table*}

\begin{table*}[]
	\begin{center}
		\resizebox{\linewidth}{!}{
			\begin{tabular}{|c|c|c|c|c|c|c|c|}
				\hline
				\multicolumn{8}{|c|}{\textbf{Appendix III: Object word table, 3/5}} \\  \hline
			    whale & flamingo & fixture & dove & blueberry & shaft & purse & waist \\ \hline	
burger & sofa & scent & tablecloth & zipper & lamb & fiber & telescope \\ \hline	
tug & downside & reptile & spire & haircut & basketball & ballet & snout \\ \hline	
bracket & hillside & booth & drone & terrain & feline & birdie & dolphin \\ \hline	
faucet & skater & skeleton & roller & dapple & hunter & rocky & crosse \\ \hline	
fungus & slipper & jungle & beef & dripping & skateboard & cane & pyramid \\ \hline	
court & blotch & toothbrush & habitat & pug & sprout & melon & sore \\ \hline	
teenager & cemetary & picnic & laundry & critic & soda & stocking & medal \\ \hline	
yolk & farmer & keepsake & speckle & stove & cafe & fiddle & barb \\ \hline	
gator & microscope & ledge & microphone & coral & trumpet & sparkler & debris \\ \hline	
gleam & drawer & wig & stairway & leafs & sunburst & wrapper & calf \\ \hline	
motor & throat & paradise & pathway & shaddow & gravel & clump & machinery \\ \hline	
teen & medicine & jellyfish & goggle & leash & snowman & weapon & pano \\ \hline	
cereal & sandal & lego & brow & sneaker & gravestone & slot & fencing \\ \hline	
hallway & mint & chopstick & shack & socket & syrup & slogan & skyscraper \\ \hline	
shovel & sponge & domino & brickwork & squash & swimmer & stool & penguin \\ \hline	
hut & typewriter & ditch & toaster & ballerina & stumper & trailer & porcelain \\ \hline	
saucer & frown & hoop & dye & quilt & utensil & portal & podium \\ \hline	
leopard & fin & burner & taxi & starfish & rooftop & poop & cottage \\ \hline	
ketchup & clay & climber & plank & boxer & opera & grabber & cab \\ \hline	
zombie & mold & bluebird & eraser & rocket & wonderland & plumage & ivy \\ \hline	
cushion & carriage & canopy & fluff & downhill & puppet & dvd & sparrow \\ \hline	
lint & thistle & golfer & coaster & closet & underwear & volcano & brass \\ \hline	
capsule & cabinet & oven & thigh & streetlight & candlelight & comb & cannon \\ \hline	
silverware & moose & robin & boardwalk & vortex & broom & tornado & workshop \\ \hline	
disc & starter & baffle & donut & theater & clover & jetty & napkin \\ \hline	
sandyp & bowling & drummer & kaleidoscope & flute & noodle & staple & reel \\ \hline	
basement & magnet & fireplace & silo & sunlit & poker & rivet & paddle \\ \hline	
vessel & rib & skintone & tigher & bristle & cattle & iceland & attire \\ \hline	
tummy & galaxy & bake & swamp & sill & vulture & gorilla & marshmallow \\ \hline	
accessory & perfume & buoy & cricket & hotpasta & muzzle & ferris & robe \\ \hline	
coach & carton & canal & boulder & wool & tatoo & thunder & mustache \\ \hline	
opener & piggy & lollipop & hairdo & donkey & watermelon & exhibition & blond \\ \hline	
sailor & alligator & daffodil & buggy & fingertip & lichen & knight & archway \\ \hline	
asphalt & mush & whistle & backpack & pylon & mic & tine & santa \\ \hline	
slat & spaghetti & client & tyre & driveway & harbor & ashtray & sushi \\ \hline	
hoof & sting & badge & contast & vodka & muckpond & lineup & cape \\ \hline	
grocery & coal & surfing & skier & rockin & kayak & forearm & cutter \\ \hline	
champagne & driftwood & elk & hound & dryer & salmon & jug & kangaroo \\ \hline	
claws & footwear & cock & maple & cabbage & oak & wildflower & goldfish \\ \hline	
slug & lobster & pub & gym & steak & blouse & jail & gutter \\ \hline	
bun & gown & vampire & groove & kiwiness & television & saddle & ash \\ \hline	
spaceship & sunray & vest & suburbia & duckie & cleaner & whit & fireman \\ \hline	
mercury & ballon & mound & railway & axe & wilderness & fender & wardrobe \\ \hline	
plaster & housing & teapot & guest & pout & duct & hen & scoop \\ \hline	
strait & toothpick & pasture & crater & bikini & gloom & symphony & dimple \\ \hline	
ferry & satellite & hopper & aluminum & butts & ape & automobile & vapor \\ \hline	
		\end{tabular}}
		\caption{Object word table}
		\label{tab:related to object words table supplementary material 3}	
	\end{center}
\end{table*}

\begin{table*}[]
	\begin{center}
		\resizebox{\linewidth}{!}{
			\begin{tabular}{|c|c|c|c|c|c|c|c|}
				\hline
				\multicolumn{8}{|c|}{\textbf{Appendix III: Object word table, 4/5}} \\  \hline
cuff & duckling & hedge & birch & rhino & meadow & raspberry & aroma \\ \hline	
bob & outing & mow & handrail & cupcake & plaid & kettle & muck \\ \hline	
lettuce & shroom & darkroom & tapestry & blaze & hippo & satin & athlete \\ \hline	
wedge & baloon & hiker & ribon & scrap & ribboner & bacon & biscuit \\ \hline	
lamppost & cacti & smog & tabletop & cobblestone & redhead & garment & pickle \\ \hline	
elevator & rifle & chef & shroud & wand & bridal & spout & filament \\ \hline	
puffin & colorcarnival & croc & fuel & pallet & rodeo & hockey & tendril \\ \hline	
headphone & monk & coastline & batman & skyscape & rodent & camping & snaffle \\ \hline	
hairline & sunbeam & lunar & suitcase & incandescent & propeller & corkscrew & yoga \\ \hline	
condom & pantie & flour & mammal & mosquito & uphill & mustard & bot \\ \hline	
wreath & dime & smoky & cradle & trophy & nun & jersey & spade \\ \hline	
pore & moustache & roadway & guitarist & spose & armpit & gourd & cobwebs \\ \hline	
resident & mallard & trampoline & llama & sentinel & mermaid & lighter & cig \\ \hline	
bathtub & hull & corridor & crate & toothpaste & prince & lmfao & diver \\ \hline	
shrimp & coyote & cocktail & talon & barbie & grafitti & toenail & buffalo \\ \hline	
piling & dough & marsh & chandelier & bullseye & hoe & waterline & vendor \\ \hline	
hamster & willow & woodpecker & putting & attic & plume & cinema & lilac \\ \hline	
circus & bait & stairwell & frisbee & complexion & trolley & pendant & iceberg \\ \hline	
gazebo & stubble & mammoth & carousel & turbine & lampshade & bovine & skys \\ \hline	
amphibian & trouser & lining & jaggy & roofline & diaper & ashe & airbrush \\ \hline	
elf & compass & paintbrush & pansy & raven & flagpole & ruffle & hail \\ \hline	
whiskey & marine & parachute & supper & chapel & jigsaw & tomb & wheelchair \\ \hline	
beagle & poodle & wick & beastie & priest & rag & cylinder & nurse \\ \hline	
mower & mike & flooring & cobble & waddy & infant & washer & slate \\ \hline	
patio & seashell & keyhole & stall & -shutterfly & pistol & lemonade & mole \\ \hline	
arena & timber & spiderweb & billow & liner & granite & hurricane & seaweed \\ \hline	
righthand & pottery & birdhouse & mascara & crude & dent & locker & eyewave \\ \hline	
tit & tress & chipmunk & gargoyle & freezer & pajama & earing & seasick \\ \hline	
linen & shawl & refrigerator & clutch & planter & doe & yacht & herb \\ \hline	
coconut & nestle & stroller & shrubbery & hammock & mop & muffin & moire \\ \hline	
ram & skylight & disco & runway & grinder & loo & handlebar & tortoise \\ \hline	
tablet & lifeguard & crescent & octopus & cattail & tounge & pepsi & fossil \\ \hline	
badger & spear & crocodile & pinecone & prairie & cloudscape & cupboard & stopper \\ \hline	
platter & packet & finch & hairstyle & coffe & hairbrush & relic & awning \\ \hline	
grime & skiing & racket & origami & sprinkler & widescreen & smokestack & citymar \\ \hline	
stapler & scrapbook & dynamite & wingtip & thermometer & digit & grease & tentacle \\ \hline	
crotch & blot & shepherd & icecream & woodland & niche & headwear & ivory \\ \hline	
parasol & lefthand & whiz & budgie & freeway & hiking & cheesecake & monopod \\ \hline	
dresser & crinkle & windscreen & chickadee & lung & classroom & cicada & supermarket \\ \hline	
eater & clothespin & tryptich & cheetah & envelop & wheelbarrow & seaside & jewelery \\ \hline	
screwdriver & meteor & broccoli & aisle & beee & dashboard & wiper & snot \\ \hline	
pew & foal & patern & tarp & cob & youngling & sfalice & checkerboard \\ \hline	
visceral & cormorant & loaf & harp & collie & dispenser & confetti & ladybird \\ \hline	
fir & gadget & pancake & bonnet & anemone & comet & withe & butcher \\ \hline	
otter & crosswalk & harbour & froggy & cinnamon & diorama & camper & botany \\ \hline	
ginger & bonsai & sweatshirt & liquor & raft & sausage & gardening & hamburger \\ \hline	
iridescence & mirage & eve & sunspot & crackle & striation & surfboard & treetop \\ \hline	
flea & snowfall & chimp & lotus & pretzel & cello & stucco & ember \\ \hline	
pistil & smoothe & suburb & locomotive & casino & boa & blender & nightshot \\ \hline	
doorknob & fairyland & cucumber & oyster & vanilla & thong & foilage & flamingos \\ \hline	
emote & shoelace & appetite & citrus & bur & pastry & matchstick & algae \\ \hline	
maid & roadside & fishnet & tram & iguana & colur & handbag & wither \\ \hline	
hourglass & sap & pussy & chore & mob & scorpion & potty & cassette \\ \hline	
christma & riveting & lingerie & paperclip & hibiscus & skiff & palace & taillight \\ \hline	
obstacle & nightscape & dropper & kicker & obscurity & chili & crutch & mite \\ \hline	
cola & paranormal & placemat & horseshoe & apature & gin & paperweight & mineral \\ \hline	
chameleon & navel & teller & barber & scarecrow & asparagus & lanscape & mansion \\ \hline	

		\end{tabular}}
		\caption{Object word table}
		\label{tab:related to object words table supplementary material 1}	
	\end{center}
\end{table*}

\begin{table*}[]
	\begin{center}
		\resizebox{\linewidth}{!}{
			\begin{tabular}{|c|c|c|c|c|c|c|c|}
				\hline
				\multicolumn{8}{|c|}{\textbf{Appendix III: Object word table, 5/5}} \\  \hline
shotgun & ranch & binocular & pomegranate & glider & zigzag & retina & sprig \\ \hline	
aspen & mortar & stork & astrophotography & riding & slab & ala & tobacco \\ \hline	
rugby & jog & assemblage & gecko & campfire & textile & dumpster & eyeliner \\ \hline	
shingle & skateboarder & ware & starbuck & beaker & glassware & patchwork & cycling \\ \hline	
netting & hump & hunk & turf & jade & cloak & campus & slime \\ \hline	
fragrance & redneck & sprite & clarinet & valve & artichoke & hoodie & pom \\ \hline	
craftsmanship & oasis & cobalt & streetpigeon & gardener & mosque & wingspan & beater \\ \hline	
blackberry & hare & osprey & covering & agave & jockey & clam & rind \\ \hline	
motorbike & egyptian & fireworks & fibre & pacifier & stag & quill & beady \\ \hline	
rocker & dogwood & cellar & totem & sleeper & overpass & scenary & gore \\ \hline	
fort & terrier & raisin & rump & eyeglass & prong & sandstone & xmas \\ \hline	
panda & vaseline & ruby & bandana & shotty & lobby & jogger & undulation \\ \hline	
glaci & goblet & pest & pong & lightpost & hurdle & inn & greenhouse \\ \hline	
stature & tricycle & bevel & starlight & manhole & flank & beaver & landscaping \\ \hline	
woodwork & streamer & headband & outhouse & avocado & swath & hydrangea & spruce \\ \hline	
skeletal & possum & memento & fowl & equine & hive & armor & palate \\ \hline	
vineyard & herfotoman & explorer & racer & nectar & colt & buddha & cobweb \\ \hline	
warehouse & undie & seafood & beatle & marathon & bison & tuna & marina \\ \hline	
ore & airshow & streaky & chamber & caffeine & knitting & kayaker & calligraphy \\ \hline	
pigtail & grove & streetlamp & cantaloupe & chessboard & avenue & throne & angling \\ \hline	
lattice & blizzard & blower & caramel & submarine & luggage & miner & eggplant \\ \hline	
doodle & ostrich & condo & firetrail & cauliflower & falcon & takin & tar \\ \hline	
thimble & grapefruit & dahlia & halogen & lampost & reindeer & mage & duckle \\ \hline	
mattress & mule & wicker & mare & squid & oar & canister & floss \\ \hline	
scowl & lass & beggar & nutty & apron & cranberry & troop & anther \\ \hline	
meerkat & sitter & courtyard & haystack & pizzazz & headgear & tart & licorice \\ \hline	
hotdog & aphid & golferdds & chilli & headdress & nog & neckline & tomatoe \\ \hline	
dandilion & orbit & pork & eggshell & vane & cockpit & woodgrain & buzzard \\ \hline	
clove & waterway & snoot & sundown & jello & trooper & bauble & vapour \\ \hline	
thunderstorm & engraving & prom & carbon & damselfly & crib & starling & husk \\ \hline	
waterscape & tinker & fluke & frilly & chainsaw & crucifix & lolly & barge \\ \hline	
ribboning & paving & orchestra & tangerine & snowball & fucus & mummy & coffin \\ \hline	
gorge & rower & waterfront & navy & sewing & bagel & honeycomb & trike \\ \hline	
nebula & eyedropper & shampoo & mountian & eyeshadow & pulp & lightroom & cellophane \\ \hline	
blackbird & ewe & garb & cutlery & bloodshot & scaffold & altar & urn \\ \hline	
dewdrop & tequila & tabby & feller & xray & whisky & spa & firefly \\ \hline	
mascot & nymph & grille & fishie & babys & cocoon & raccoon & peppercorn \\ \hline	
retriever & sunflare & brook & lemur & farmhouse & mitten & cartridge & bomber \\ \hline	
chestnut & boater & chillin & scaffolding & goatee & welder & terrace & froggie \\ \hline	
petunia & shrine & mistiness & furnace & fisher & snowstorm & gingerbread & seedling \\ \hline	
doily & ambulance & buttock & sweety & snowboard & teardrop & kilt & gravy \\ \hline	
lumber & windowsill & pail & sled & macaw & walnut & gerbera & apropos \\ \hline	
floorboard & fume & sunburn & sod & doughnut & moor & wineglass & elder \\ \hline	
swimsuit & undieyatch & marshmellow & bicyclist & flask & motel & trout & silkiness \\ \hline	
marquee & tassel & banister & storefront & spinach & stigma & lioness & celery \\ \hline	
sideway & rollercoaster & jeeper & lighting- & arsenal & crocus & embroidery & coating \\ \hline	
baseboard & duster & geyser & pulley & lug & wharf & scooter97 & puck \\ \hline	
almond & tshirt & hardwood & ferret & reef & carpenter & lifesaver & rink \\ \hline	
catnip & graffitti & gossamer & monolith & redeye & tusk & whirlpool & sewer \\ \hline	
bleacher & cedar & cargo & yogurt & beret & eyes & tarmac & stoplight \\ \hline	
drizzle & serpentine & opal & popsicle & mining & placer & chihuahua & plumber \\ \hline	
trashcan & clipper & barnacle & antelope & gumball & rafter & tuff & silverfoxx \\ \hline	
quartz & carp & orchard & snowboarder & gauze & cavern & plexiglass & clematis \\ \hline	
farmland & saber & fencepost & lodge & sprocket & easel & peony & gamble \\ \hline	
cove & puss & chickene & drumstick & digger & creme & roadsign & cuppa \\ \hline	
bartender & sandpaper & milkweed & volleyball & hornet & barley & rum & petrol \\ \hline	
ale & draft & grizzly & doorframe & ponytail & bandage & siren & briefcase \\ \hline	
basin & stylist & scoreboard & shipwreck & statuary & fenceline & rash & loft \\ \hline	
bookcase & bandaid & scallop & beet & plier & vinegar & parchment & cowgirl \\ \hline	
blindfold & trinket & dorm & burlap & halter & jewlery & bookshelf & spatula \\ \hline	
paver &   &   &   &   &   &   &   \\ \hline

		\end{tabular}}
		\caption{Object word table}
		\label{tab:related to object words table supplementary material 2}	
	\end{center}
\end{table*}

\end{document}